\documentclass[11pt, twocolumn]{article}

\usepackage[T1]{fontenc}
\usepackage{lmodern}
\usepackage{enumitem}
\usepackage[margin=1in]{geometry}

\usepackage{authblk}

\usepackage[colorlinks=true, linkcolor=blue, citecolor=blue, urlcolor=blue]{hyperref}

\newif\ifbuildpreprint
\buildpreprinttrue  

\usepackage{xfrac}
\usepackage{graphicx}
\usepackage{booktabs}
\usepackage{listings}
\usepackage{multirow}
\usepackage{tabularx}
\usepackage{threeparttable}
\usepackage{bm}
\usepackage{upgreek}
\usepackage{mathtools} 
\usepackage{algorithm}
\usepackage{algpseudocode}
\usepackage{txfonts}
\usepackage{dutchcal}
\usepackage{microtype} 
\usepackage{setspace}
\usepackage{cleveref}
\usepackage{lineno}
\ifbuildpreprint
\else
  \linenumbers
\fi

\DeclareMathAlphabet{\mathcalorig}{OMS}{cmsy}{m}{n}

\newcommand{\ve}[1]{\bm{#1}}  
\newcommand{\vs}[1]{\mathcal{#1}}  

\title{\textbf{Task- and dataset-specific information in protein language models}}

\author[1,2]{Roman Joeres\thanks{R.J. and I.S. contributed equally to this work.}}
\author[*1,2]{Ilya Senatorov}
\author[1,2,3]{Anastasia Kolchina}
\author[2,3,4]{Dietrich Klakow}
\author[1,2,4,5]{Olga V. Kalinina\thanks{Corresponding author: olga.kalinina@helmholtz-hips.de}}

\affil[1]{Drug Bioinformatics, Helmholtz Institute for Pharmaceutical Research Saarland, 66123 Saarbruecken, Germany}
\affil[2]{Center for Bioinformatics, Saarland University, 66123 Saarbruecken, Germany}
\affil[3]{Saarland Informatics Campus, Saarland University, 66123 Saarbruecken, Germany}
\affil[4]{Pharma Science Hub, 66123 Saarbruecken, Germany}
\affil[5]{Medical Faculty, University Hospital Saarland, 66421 Homburg, Germany}

\date{}

\begin{document}

\onecolumn 

\maketitle

\begin{abstract}
Protein language models (PLMs) have transferred the latest advances from natural language processing to computational biology.
These models, trained on large corpora of protein sequence data, are widely used to translate amino acid sequences into latent-space embeddings, ready for use in diverse downstream tasks (DTs).
By consensus, embeddings from the models' last layers are used, while the models' internal behavior remains poorly understood.

We analyzed 13 PLMs across 15 DTs and 9 datasets to assess the value of embeddings from intermediate PLM layers.
We trained probe models on embeddings from each layer, compared their performance, and showed that the last layers of PLMs rarely produced embeddings that led to the best results on downstream tasks.
Furthermore, we identified a connection between how models learn a certain DT and the similarity between that DT and the pre-training objective.
For example, for residue-level downstream tasks, we observed a steady increase in performance across almost all PLM layers, which we attributed to their similarity to most PLMs' pre-training objectives.

To allow the community to capitalize on our findings, we provide \verb+PLMSommelier+, a Python package that automatically identifies the best PLM layer for a given DT with $\sim$98\% accuracy and creates a truncated model using only the early layers up to the best-performing layer. This will help users save time and memory during inference and yield better predictive performance.
\end{abstract}

\vspace{1em}
\vspace{1em}

\clearpage 
\onehalfspacing

\section{Introduction}

Since the release of ESM-1 in 2019~\cite{rives2021biological}, the first transformer-based protein language model (PLM), their popularity has steadily risen.
PLMs are often used as crucial components in feature engineering processes for predicting protein properties and functions, such as enzymatic activity, binding-site identification, and antibody design~\cite{capela2025comparative, teukam2024language, shuai2023iglm, leclercq2025protein}.
However, it is still not well understood how PLMs represent proteins internally, what information they extract, where across layers they store it, and whether this varies by protein type, downstream task, or dataset. 

Following deep learning (DL) theory, DL models compose abstractions within their internal representations, also called \emph{layers}, incrementally building more informative embeddings from shallow layers near the input to deep layers near the full depth, with the last, deepest layer being ``the best'' in a DL model.
This is supported by Alain and Bengio~\cite{alain2016understanding}, who used \textit{linear classifier probes} to investigate how informative each layer of a deep neural network is, with respect to a downstream application, also called a \emph{downstream task} (DT).
For the computer vision model ResNet-50~\cite{he2016deep}, they found that the prediction error decreases steadily with each deeper layer of the network.
They showed that the latent spaces of deeper layers become more linearly separable, which is intuitive, since the model is trained end-to-end on the ImageNet~\cite{deng2009imagenet} dataset and the last layer is a linear classifier over the final embedding.
Therefore, the model naturally strives to improve linear separability with each layer.
Similarly, the last layer of a PLM is assumed to be the most informative and is therefore usually used to provide embeddings for DTs.

But large language models, including PLMs, are trained differently: they usually undergo self-supervised pre-training to gain a general understanding of the data before being applied to DTs.
For example, ESM-1~\cite{rives2021biological}, as well as other models in the ESM family, is a masked language model (MLM) for the ``protein language'': during pre-training, it predicts masked amino acids in a protein sequence. These generalist models are then fine-tuned or used to generate features for prediction heads on DTs with different objectives~\cite{devlin2019bert}.
For PLMs, this leads to a scale mismatch: pre-training typically operates locally at the token level (individual amino acids), while many downstream tasks require global, sequence-level predictions (whole proteins).
This, in turn, requires researchers to use various pooling methods, which most PLMs are not originally trained for.
Therefore, the pre-training task improves understanding at the amino acid level, but not necessarily at the whole-protein level.
We assumed that deeper PLM layers become better at the pre-training task and aimed to study how intermediate layers perform on various downstream tasks.

Indeed, evidence suggests that intermediate PLM layers may contain more valuable information for DTs than the last layer.
For example, Kumar and Jha showed that using the mean of representations from mid-to-final layers' embeddings yields a more informative protein embedding than the final layer alone in a kinase function prediction DT~\cite{kumar2025layer}.
Along the same lines, Vig et al. showed that the shallow layers of PLMs learn simpler biophysical properties, such as amino acid features and secondary structures, while deeper layers capture more complex properties, such as binding sites and contact maps~\cite{vig2020bertology}.
This aligns with the behavior of neural networks in computer vision, where kernels in shallow layers learn simple concepts like lines and curves, while kernels in deeper layers learn more complex features, such as faces and objects~\cite{zeiler2014visualizing}.
However, the lack of a systematic analysis across many PLMs and DTs does not yet allow us to draw conclusive lessons from these observations.

In this study, we investigate this phenomenon by training probes on a variety of PLMs and DTs. We analyze 13 models from five model families across four LLM architectures and 9 datasets, spanning 15 DTs, to develop a well-founded understanding of PLM layer behavior. 
We show that model performance varies across internal layers' embeddings, the performance trends differ across DTs, and propose an explanation based on the DTs' objectives and data.
We converted our findings into \verb+PLMSommelier+, a Python package that calculates the best-performing layer of a PLM and provides the user with a truncated PLM.
It improves performance, reduces inference runtime, and decreases the size of models stored locally.

\begin{figure}[t!]
    \centering
    \includegraphics[width=\textwidth]{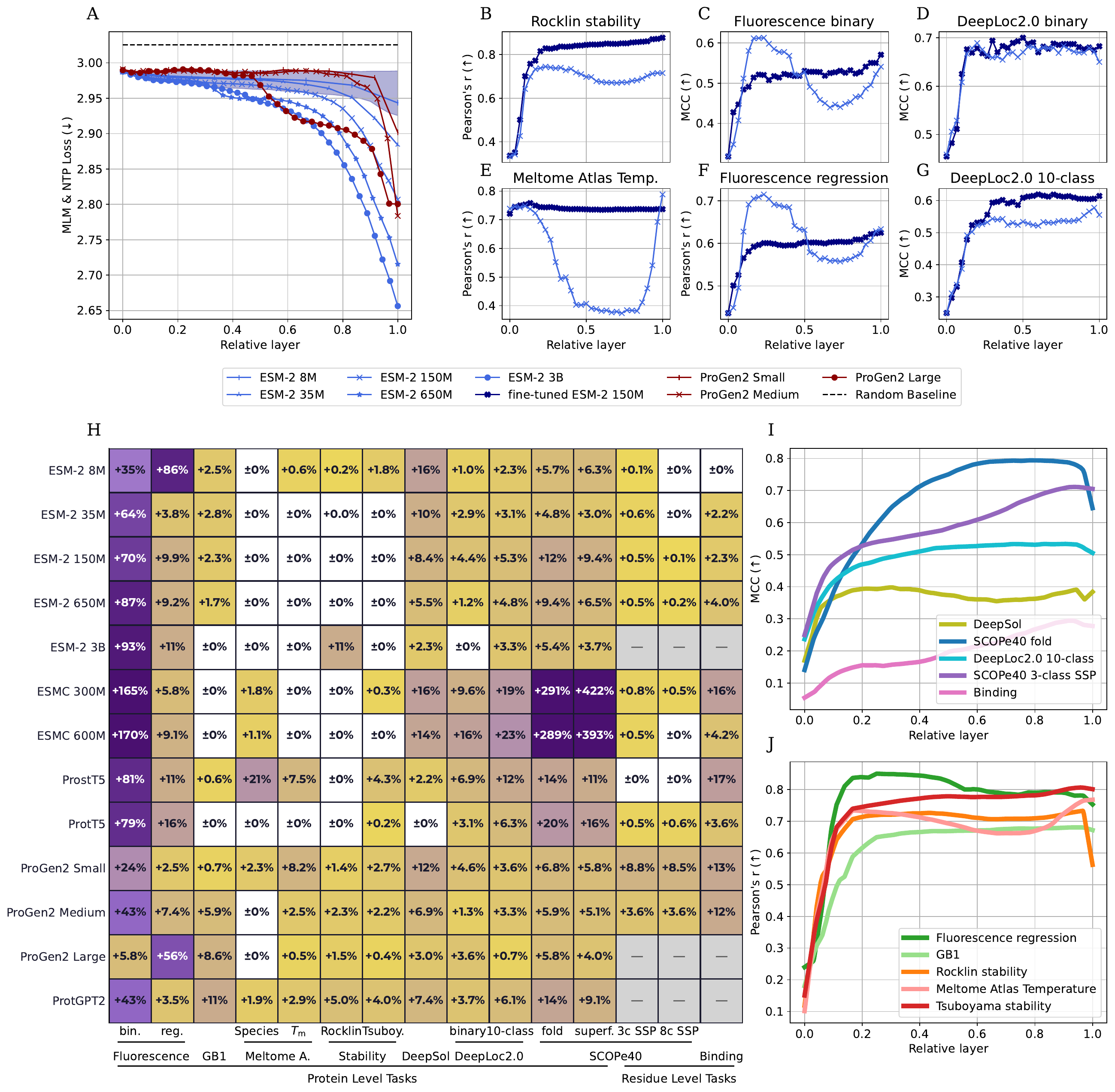}
    \caption{\textbf{Performance of internal PLM layers on diverse DTs.} \textbf{A} Native PLMs improve in each layer on their pre-training objective. This is true for both the MLM objective (ESM-2 models) and the autoregressive objective (ProGen2 models). Fine-tuning the PLM worsens the performance on the pre-training objective (shaded area for fine-tuned ESM-2 150M). The dashed black line shows the performance of a random baseline, averaged over 5 runs. \textbf{B-G} fine-tuning improves each layer over the previous one in predictiveness toward the fine-tuning objective. \textbf{H}: Relative improvement of the best layer of each PLM over the last layer on all DTs. ESM2-3B and ProGen2-Large required too much storage to compute residue-wise embeddings; ProtGPT2 has no per-residue embeddings, thus having no results for those tasks. $\pm0\%$ improvement implies the last layer was the best one. \textbf{I} and \textbf{J}: Average performance for each dataset across all models for classification (I) and regression (J) tasks.}
    \label{fig:perf}
\end{figure}

\section{Results}

We used 13 protein language models (PLMs) to compute protein-level embeddings for eight datasets~\cite{sarkisyan2016local, olson2014comprehensive, rocklin2017global, tsuboyama2023mega, jarzab2020meltome, khurana2018deepsol, thumuluri2022deeploc, chandonia2022scope} and residue embeddings for two~\cite{chandonia2022scope, littmann2021protein} (SCOPe40 was used in both cases). These datasets correspond to the following 15 DTs: fluorescence intensity regression and active/inactive classification on the fluorescence dataset~\cite{sarkisyan2016local}, mutational effect regression on the GB1 dataset~\cite{olson2014comprehensive}, stability score regression from the Rocklin and Tsuboyama stability datasets~\cite{rocklin2017global, tsuboyama2023mega}, melting temperature $T_\text{m}$ regression and species classification from the Meltome Atlas~\cite{jarzab2020meltome}, 10-class classification of proteins into subcellular locations and binary membrane protein classification on DeepLoc2.0~\cite{thumuluri2022deeploc}, binary solubility classification from DeepSol~\cite{khurana2018deepsol}, fold and superfamily classification on proteins as well as 3-class and 8-class secondary structure prediction on residues from SCOPe40 2.08~\cite{chandonia2022scope}, secondary structure elements were assigned using the DSSP algorithm~\cite{kabsch1983dictionary}, and binary classification of residues w.r.t. small-molecule interactions from Littmann et al.~\cite{littmann2021protein} (see \Cref{sec:datasets} for more details). 

The 13 PLMs comprise 5 ESM-2 models~\cite{lin2023evolutionary} (ESM-2 8M, ESM-2 35M, ESM-2 150M, ESM-2 650M, and ESM-2 3B), ESMC-300m and ESMC-600m~\cite{esm2024esmcambrian}, ProtT5~\cite{elnaggar2021prottrans}, ProstT5~\cite{heinzinger2024bilingual}, ProGen2-small, ProGen2-medium, and ProGen2-large~\cite{nijkamp2023progen2}, as well as ProtGPT2~\cite{ferruz2022protgpt2}. The first 9 of them are trained bidirectionally; the last 4 are trained with an autoregressive loss, and ProtGPT2 additionally uses a BPE tokenizer for subword tokenization~\cite{gage1994new} (see \Cref{sec:plm} for details).
These PLMs are based on four different language model architectures. The original \textit{transformer} architecture embeds the self-attention mechanism into an encoder-decoder architecture, originally designed for language translation tasks~\cite{vaswani2017attention}. \textit{BERT} extracted the encoder of the transformer architecture to use it bidirectionally, optimizing it for sequence- and token-level understanding through feature extraction or task-specific fine-tuning~\cite{devlin2019bert}. The \textit{T5} architecture returns to the full encoder-decoder architecture and is designed to unify all NLP tasks by framing them as ``text-to-text'' tasks~\cite{raffel2020exploring}. Finally, the \textit{generative pre-trained transformer}, GPT, is a unidirectionally trained architecture for token generation~\cite{radford2018improving}.

For each PLM-DT pair, we fit two probes: for classification DTs, a logistic regressor and a $k$-NN classifier; for regression datasets, a linear regressor and a $k$-NN regressor. These probes capture both linear and non-linear features in the embedding spaces. For the $k$-NN probes, we chose $k=10$.
We selected these probes because they have few hyperparameters and can be fit to the data rather than trained for prolonged periods of time.
After fitting the probes to the training split, we evaluate and report their performance on the validation splits (Matthews correlation coefficients for classification and Pearson correlation coefficients for regression).

\subsection{End-to-end vs. piecewise training}\label{sec:e2e}
As illustrated by ResNet-50~\cite{he2016deep}, when a model is trained end-to-end on a specific task, each layer contributes to solving that task, and layers sequentially become better at predicting that task.
We hypothesize that when operating within the paradigm of foundational models, i.e., first pre-training a model, then training a downstream prediction head on its embeddings, layers might not improve sequentially in predicting the downstream task.
To test this, we conduct three experiments: (i) probe the layers of PLMs on their pre-training objective, (ii) fine-tune PLMs on DTs and probe the layers on these DTs again, and (iii) take those fine-tuned PLMs and probe their layers on the pre-training objective again.

We conducted the first experiment on the ESM-2 model family with the MLM loss and on the ProGen2 model family with the next token prediction (NTP) loss.
To create the MLM loss dataset, we sampled 10,000 sequences uniformly from the SCOPe40 dataset and masked 15\% of the residues. For the NTP loss dataset, we uniformly sampled a position in each of the 10,000 sequences, truncated the sequence at that position, and used the next token as the label. Then, we embedded the amino acids using the ESM-2 and ProGen2 models and fitted a linear probe to predict the amino acid type from the embeddings of the masked residues and the next token from the last residue embedding of the truncated sequences, respectively.
Pre-training losses for ESM-2 and ProGen2 decrease almost monotonically with depth (\Cref{fig:perf}A), showing that PLMs steadily optimize their training objective at deeper layers.
For the second experiment, we fine-tuned the ESM-2 150M model on six DTs (AdamW with a learning rate of 1e−4, batch size 32, and a maximum of 100 epochs, with early stopping after 10 epochs without improvement in validation loss and selection of the best-validation checkpoint), trained linear probes on the fine-tuned ESM-2 layers, and compared their performance with linear probes trained on the original ESM-2 models (\Cref{fig:perf}B-G). Fine-tuning shows a clear effect, and the performance of the fine-tuned models generally increases monotonically, indicating that the native PLM's non-monotonic behavior stems from a mismatch between its pre-training objective and the DT.
The third experiment recalculated the MLM loss for each layer of the fine-tuned ESM-2 150M models. Using the MLM loss dataset from the first experiment, we probed the layers of the six fine-tuned models (shaded area in \Cref{fig:perf}A).
Fine-tuning clearly shifts each layer's focus away from the pre-training objective and toward the DT.

These experiments show that a PLM that is pre-trained on a residue-level objective improves in each layer with respect to residue-level DTs, with small drop-offs in performance in the last layer (\Cref{fig:ablation}D and \Cref{fig:perf}A). But when it is applied to protein-level tasks, the last layer is not the best (\Cref{fig:perf}B-G). When the PLM is then end-to-end fine-tuned on a protein-level DT, it improves in each layer with respect to that protein-level DT (\Cref{fig:perf}B-G), but drops in residue-level performance (\Cref{fig:perf}A). This observation is supported by the fact that performance steadily increases for residue-level tasks, where the objective by design aligns better with the pre-training objective that is also residue-level (\Cref{fig:ablation}D and Supplementary Figure 2 M, N, and O).

\begin{table*}[t!]
    \centering
    \caption{Comparison of the performance of linear probes on the best layers of all PLMs and all tasks. Regression tasks (marked $^\dagger$) report Pearson correlation (PCC); all other tasks report Matthews correlation coefficient (MCC). ESM2-3B and ProGen2-Large required too much storage to compute residue-wise embeddings; ProtGPT2 has no per-residue embeddings, thus having no results for those tasks (gray tiles).}
    \label{tab:best_values}
    \resizebox{\textwidth}{!}{%
    \begin{tabular}{lcccccccccccc|ccc}
        & \multicolumn{12}{c|}{\textsc{protein-level tasks}} & \multicolumn{3}{c}{\textsc{residue-level tasks}}\\
        \textbf{Models} & \multicolumn{2}{c}{\textbf{Fluorescence}} & \textbf{GB1}$^\dagger$ & \multicolumn{2}{c}{\textbf{Meltome Atlas}} & \multicolumn{2}{c}{\textbf{Stability}} & \textbf{DeepSol} & \multicolumn{2}{c}{\textbf{DeepLoc2.0}} & \multicolumn{2}{c|}{\textbf{SCOPe40}} & \multicolumn{2}{c}{\textbf{SSP}} & \textbf{Binding} \\
        & Bin. & Reg.$^\dagger$ &  & $T_\text{m}^\dagger$ & Species & Rocklin$^\dagger$ & Tsubo.$^\dagger$ &  & Bin. & 10c & Fold & SF. & 3c & 8c & \\ \midrule
        ESM-2 8M & 0.546 & 0.749 & 0.580 & 0.611 & 0.708 & 0.707 & 0.666 & 0.360 & 0.641 & 0.442 & 0.707 & 0.731 & 0.635 & 0.516 & 0.149 \\
        ESM-2 35M & 0.624 & 0.807 & 0.630 & 0.632 & 0.739 & 0.724 & 0.752 & 0.377 & 0.673 & 0.487 & 0.787 & 0.794 & 0.700 & 0.574 & 0.241 \\
        ESM-2 150M & 0.742 & 0.837 & 0.652 & 0.699 & 0.769 & 0.734 & 0.813 & 0.410 & 0.701 & 0.534 & 0.825 & 0.821 & 0.746 & 0.631 & 0.339 \\
        ESM-2 650M & 0.815 & 0.883 & 0.718 & 0.761 & 0.795 & 0.747 & 0.845 & 0.413 & 0.714 & 0.583 & 0.851 & 0.835 & 0.780 & 0.673 & 0.332 \\
        ESM-2 3B & \textbf{0.888} & 0.881 & \textbf{0.786} & 0.789 & 0.798 & \textbf{0.769} & 0.859 & 0.433 & \textbf{0.732} & \textbf{0.621} & \textbf{0.854} & \textbf{0.837} & -- & -- & -- \\
        ESMC 300M & 0.784 & 0.870 & 0.685 & 0.766 & 0.837 & 0.766 & 0.859 & 0.423 & 0.700 & 0.557 & 0.846 & 0.829 & 0.780 & 0.668 & 0.331 \\
        ESMC 600M & 0.837 & 0.881 & 0.705 & 0.792 & \textbf{0.853} & 0.768 & \textbf{0.871} & 0.425 & 0.722 & 0.577 & 0.843 & 0.837 & 0.791 & 0.683 & 0.350 \\
        ProstT5 & 0.816 & 0.876 & 0.714 & 0.643 & 0.760 & 0.748 & 0.841 & 0.421 & 0.705 & 0.546 & 0.850 & 0.831 & \textbf{0.821} & \textbf{0.733} & \textbf{0.417} \\
        ProtT5 & 0.821 & 0.875 & 0.695 & \textbf{0.835} & 0.849 & 0.761 & 0.852 & 0.430 & 0.716 & 0.577 & 0.843 & 0.832 & 0.787 & 0.678 & 0.375 \\
        ProGen2 Small & 0.668 & 0.875 & 0.661 & 0.589 & 0.748 & 0.727 & 0.759 & 0.390 & 0.658 & 0.479 & 0.705 & 0.719 & 0.513 & 0.416 & 0.206 \\
        ProGen2 Medium & 0.739 & \textbf{0.886} & 0.714 & 0.724 & 0.774 & 0.750 & 0.810 & 0.419 & 0.676 & 0.529 & 0.761 & 0.769 & 0.588 & 0.480 & 0.242 \\
        ProGen2 Large & 0.842 & 0.880 & 0.763 & 0.754 & 0.766 & 0.755 & 0.835 & \textbf{0.445} & 0.672 & 0.563 & 0.750 & 0.758 & -- & -- & -- \\
        ProtGPT2 & 0.727 & 0.860 & 0.672 & 0.684 & 0.746 & 0.741 & 0.775 & 0.425 & 0.672 & 0.531 & 0.766 & 0.775 & -- & -- & -- \\ \bottomrule
    \end{tabular}%
    }
\end{table*}

\subsection{The deepest layer is almost never the best choice for use in DTs}\label{sec:layers}
\begin{figure*}[t!]
    \centering
    \includegraphics[width=\textwidth]{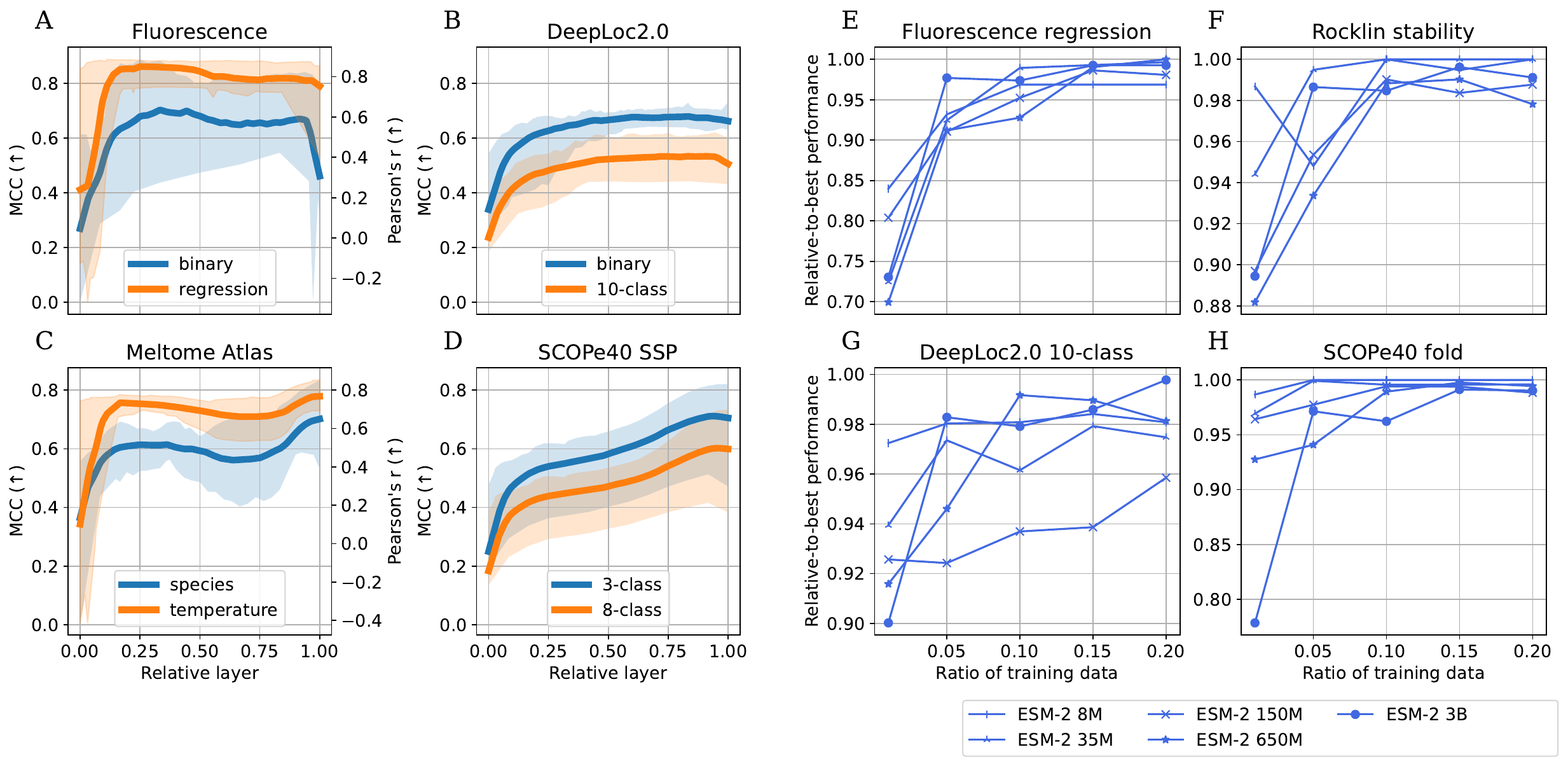}
    \caption{\textbf{Ablation studies.} \textbf{A-D}: Pairwise comparisons of two tasks for one dataset, showing that model performance patterns do not change across DTs within the same model and dataset. \textbf{E-H}: Subsampling of the DT training dataset shows that relatively little data is sufficient to find a PLM layer that achieves close-to-optimal performance. We ran these experiments over three seeds; in each, the same layer was identified as ``the best'', showing the robustness of testing on subsets.}
    \label{fig:ablation}
\end{figure*}

We assessed the performance of PLMs using linear probes (\Cref{fig:perf}) and $k$-NN probes (Supplementary Figure 1), and only in a fraction of the 186 computed cases (20.96\% for linear probes, 16.13\% for $k$-NN probes, \Cref{fig:perf}H, Supplementary Figure 1A) did we find that the embeddings of the deepest layer yielded the best performance in the corresponding DT.
We did not calculate residue-level embeddings from ESM2-3B and ProGen2 Large, because these models yield embeddings that are too large to store (\Cref{tab:plms}). Because of subword tokenization, ProtGPT2 has no per-residue embeddings and thus no results for those tasks.
For most PLMs and protein-level DTs, performance increases dramatically over the first few layers, peaks between the 10th and 90th percentiles of model depth, and then declines in the deepest layers (\Cref{fig:perf}I and J and Supplementary Figures 1B and C).
Interestingly, this differs for the residue-level DTs: there, most models show a steady increase in performance with each deeper layer (\Cref{fig:ablation}D and Supplementary Figures 2M, N, and O).
Valeriani et al. connect this phenomenon to the pre-training design, where the deepest layer predicts amino acid types and is evaluated in the MLM loss.
Therefore, PLMs have to reshape their internal representation of the protein to best distinguish amino acids~\cite{valeriani2023geometry}.
This behavior is consistent across different PLMs, datasets, and DTs.
ProGen2 Small, ProGen2 Medium, and ProtGPT2 show a smaller drop-off in the deepest layers than the MLM models, whereas ProGen2 Large behaves like the MLM models (Supplementary Figure 3).
The PLM-DT performance curves of linear and $k$-NN probes are highly similar, as shown by the median distance between their reported best layer indices of 2 (Supplementary Figure 2).

Additionally, performance across layers is largely insensitive to factors such as data split, objective, and task difficulty for a given dataset.
We demonstrate this on four datasets (fluorescence, DeepLoc2.0, Meltome Atlas, and secondary structure prediction on SCOPe40) by comparing two tasks for each that differ in split, objective, and perceived difficulty based on their intrinsic label hierarchy (\Cref{fig:ablation}A-D).
For the fluorescence dataset, the tasks differ in the objective (a classification task into active and inactive fluorescing proteins and a regression task predicting the raw log-scaled fluorescence intensity values).
We treat regression as more difficult than classification on the same dataset.
For the DeepLoc2.0 dataset, we consider binary (membrane or non-membrane) and 10-class (based on subcellular localization information) classification tasks.
Here, we consider a classification task at a lower hierarchical level (10 sub-classes rather than 2 larger classes) to be more difficult.
The Meltome Atlas offers two tasks: $T_\text{m}$ regression and species classification, which are closely related, as Lopez et al.~\cite{lopez2025supervised} show.
Similar to DeepLoc2.0, for secondary structure prediction (SSP), we consider three- and eight-class classification tasks at residue level.
The models' performance behavior for all four pairs of tasks is similar across all layers: three out of four mean Pearson correlation coefficients between the performance curves of the individual models for the datasets are $\sim$0.85 or above (fluorescence: $0.681 \pm 0.241$, DeepLoc2.0: $0.978 \pm 0.024$, Meltome Atlas: $0.847 \pm 0.103$, SCOPe40 SSP: $0.996 \pm 0.003$), indicating that the very same layers of the corresponding PLMs produce embeddings yielding highly similar performance in both task variants in all four datasets, and this is consistent across different splits, objectives, and difficulties.
We find the same for $k$-NN probes; the mean per-model Pearson correlations are above 0.9 (Supplementary Figure 5).

From a practical perspective, we wanted to determine how quickly one can identify the best PLM layer for a given DT.
To investigate this, we uniformly downsampled the full dataset and retained the data splits for all datasets except SCOPe40, for which we computed stratified splits using DataSAIL~\cite{joeres2025data} to ensure each split included members of every class.
We then probed the downsampled datasets with the ESM-2 models and found that only 15\% of the data is sufficient to consistently identify a layer achieving at least 95\% of the best performance. 
Additionally, we observed that larger models are more robust to sparse data in the DTs (\Cref{fig:ablation}E-H and Supplementary Figure 6 for all datasets).

\subsection{PLMSommelier}

\begin{figure*}[t!]
    \centering
    \includegraphics[width=\textwidth]{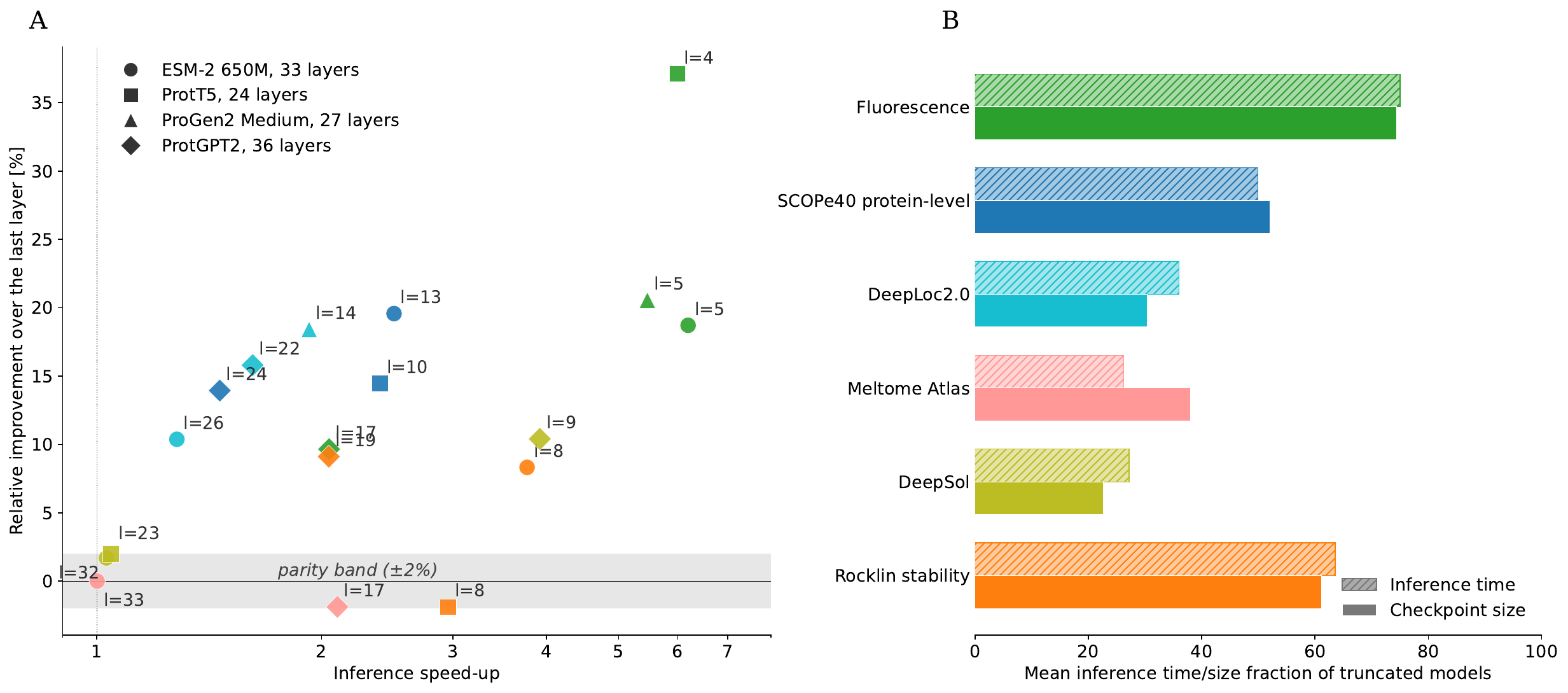}
    \caption{\textbf{PLMSommelier overview.} \textbf{A}: Improvement of the best layer vs the speed increase; the layer picked is denoted with $l=X$ \textbf{B} Speed increase and disk space saved calculated over 4 models (ESM-650M, Prot-T5, ProGen2-medium, ProtGPT2), averaged over each of the presented datasets.}
    \label{fig:benchmark_fig}
\end{figure*}

To allow the community to capitalize on our findings, we created a Python package titled \verb+PLMSommelier+.
It efficiently identifies the best layer for each DT-PLM pair, truncates the PLM by removing the layers after the selected one, and returns the truncated version to the user.
The user can then easily use the truncated model to compute embeddings for the full downstream dataset or fine-tune it.
Using truncated models not only improves performance but also saves compute and disk space.
As evidenced by \Cref{fig:ablation}E-H, only a fraction of the data is enough to identify a close-to-the-best layer, so by default \verb+PLMSommelier+ samples 15\% of the dataset.
Then it computes embeddings for all sampled points from each layer and trains a simple probe (linear/logistic regression, $k$-NN, or MLP).
Afterward, the user receives a truncated version of the corresponding PLM from the first layer up to the best-performing layer.
If multiple layers perform close to the best (within 2\%), by default, \verb+PLMSommelier+ chooses the shallowest layer.
For datasets with early peaks, such as fluorescence and stability, these truncated PLMs provide up to 6x speed-up (\Cref{fig:benchmark_fig}A). 
Truncated models also outperform last-layer-based models by up to 37\%, while the model checkpoint size decreases by 42\% on average.

As evidenced by \Cref{fig:benchmark_fig}B, some datasets benefit more from using PLMSommelier than others.
Of course, the most impressive improvements in both speed and performance are observed on tasks in which shallow layers are the best (fluorescence and Rocklin stability).
However, even for DTs like SCOPe40 fold recognition or DeepSol, for which best-performing layers are relatively deep, we observe up to 2x inference speed-up (sequences per second), which translates to over 30\% faster processing of the whole dataset.
This means that it is almost always faster to truncate the model using \verb+PLMSommelier+ first, before training a prediction head for a downstream task.

\begin{figure*}[t!]
    \centering
    \includegraphics[width=\textwidth]{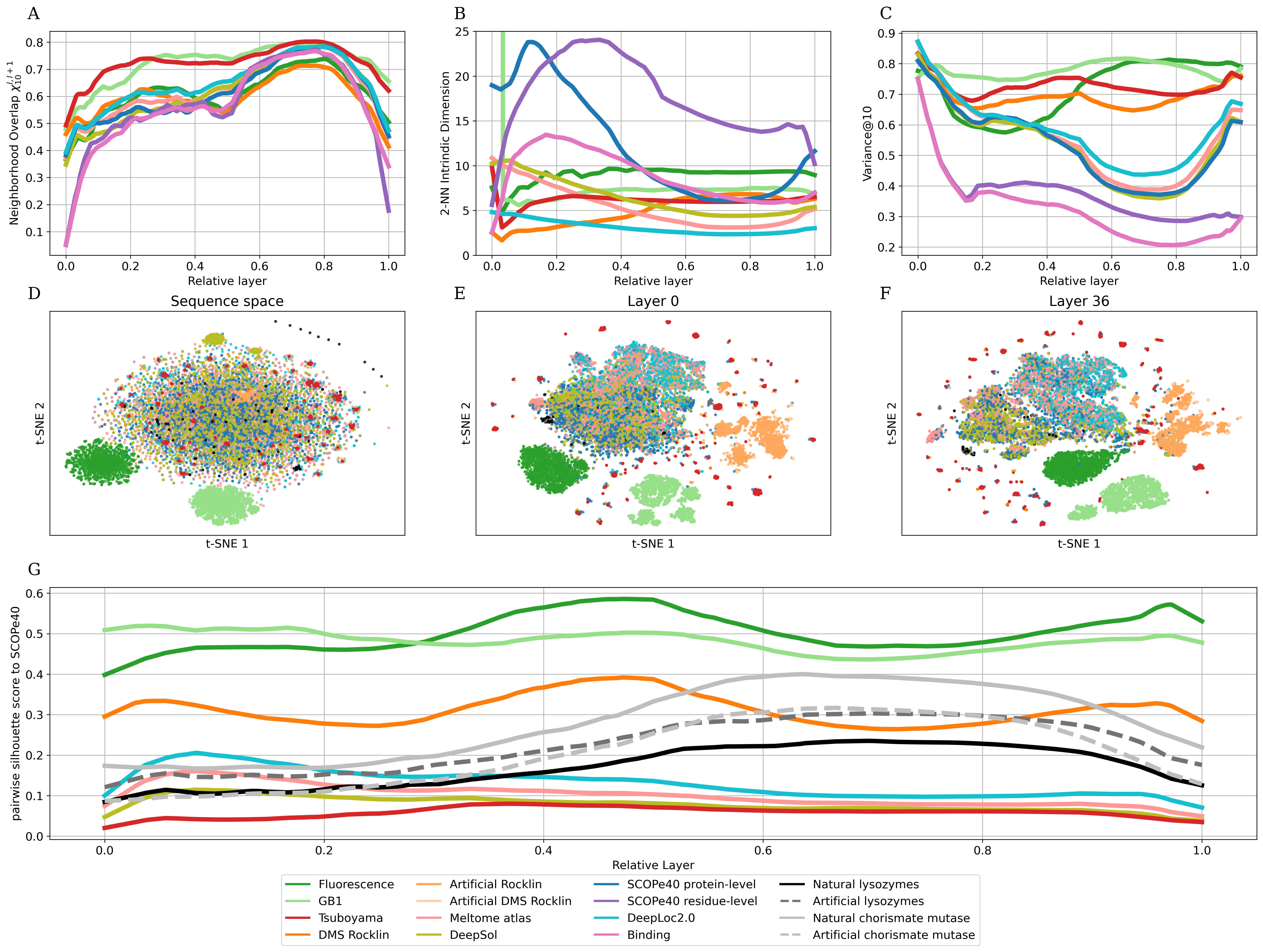}
    \caption{\textbf{Latent space analysis.} \textbf{A} Neighborhood overlap between consecutive layers of the PLMs \textbf{B} Intrinsic dimension approximated using the 2-NN method \textbf{C} Variance explained by the first 10 principal components of a latent space PCA. The Methods section explains all metrics in more detail. \textbf{D} t-SNE visualization of neighborhoods in the sequence space. Sequence identities are calculated using MMseqs2~\cite{steinegger2017mmseqs2}. \textbf{E} Visualization of the datasets through tokenization by the ESMC-600m model. \textbf{F} After the last layer of ESMC-600m. \textbf{G} Pairwise silhouette scores between the datasets and SCOPe40 through the depth of the PLMs. All lines are averaged over the 13 PLMs.}
    \label{fig:latent_sp}
\end{figure*}

\begin{figure*}[t!]
    \centering
    \includegraphics[width=\textwidth]{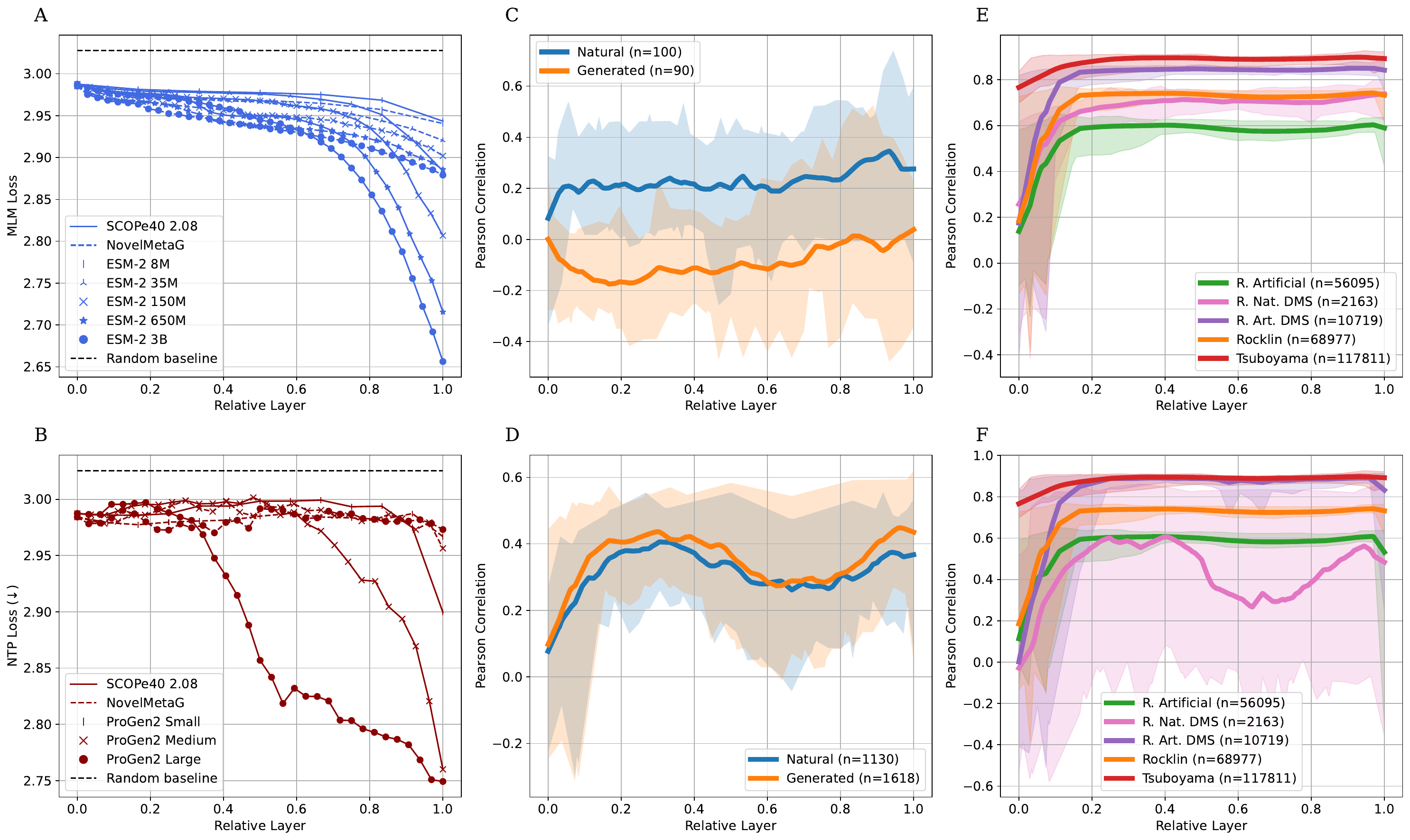}
    \caption{\textbf{Possible explanations for layerwise behavior.} \textbf{A} and \textbf{B} Comparison of the MLM and NTP loss of ESM-2 and ProGen2 models, respectively, on pre-training-included SCOPe40 sequences (solid lines) and NovelMetaG sequences with $<30\%$ sequence identity to any sequence in UniProt (dashed). The dashed black lines show the performance of a random baseline, averaged over 5 runs. \textbf{C} Comparison of predicted activity of natural and ProGen-generated lysozymes. \textbf{D} Comparison of the predicted relative enrichments between natural bmDCA-generated chorismate mutases. \textbf{E} Disentangling performance contribution to the individual parts of the Rocklin stability dataset with comparison to the performance on the full Rocklin and Tsuboyama datasets. \textbf{F} Comparison of probes trained individually on the three parts of the Rocklin stability dataset with comparison to the performance on the full Rocklin and Tsuboyama datasets. \textbf{C-F} The solid lines show the mean performance over all 13 PLMs with linear regression probes, the shaded areas show the minimum and maximum performance.}
    \label{fig:explanations}
\end{figure*}

\subsection{Topology of the latent spaces does not explain the PLMs' performance across layers}

In \Cref{sec:e2e}, we show that the alignment of the DT with the pre-training task plays a major role in shaping the layer-performance curves. However, this explains only the consistent improvement in residue-level DTs; for other DTs, the pre-training objective seems equally misaligned with the DT, yet the layer-wise performance curves do not improve monotonically and differ between datasets. We therefore investigated whether alignment between the pre-training data and the downstream data, or the structure of the downstream data itself, may also play a role.

First, we investigated how stable the subspace spanned by embeddings of a particular downstream dataset is by computing neighborhood overlaps for consecutive layers, subspace intrinsic dimension, and variance explained by the first 10 principal components (\Cref{fig:latent_sp}A-C).
The neighborhood overlap, i.e., the fraction of shared neighbors among the closest 10 neighbors of each data point in the consecutive layers, is high for all datasets, except in the shallow and very deep layers (which can be explained by the effects of the MLM objective~\cite{valeriani2023geometry}), showing that the latent space stays stable.
The intrinsic dimension, i.e., the number of independent variables needed to describe the embedding subspace, grows in the shallow layers for the SCOPe40 and binding datasets but stays under 10 for most datasets and layers (per-model curves in Supplementary Figures 7 and 8). The explained variance @10 is highest for the two DMS and two stability datasets, and decreases substantially for other datasets in the deeper half of the models.

Next, we compared the distribution of the data points in the input sequence space (using similarities calculated with MMseqs2~\cite{steinegger2017mmseqs2}) and latent spaces for a representative PLM (\Cref{fig:latent_sp}D-F). The DMS datasets, fluorescence, and GB1 are clearly separated from the rest in all spaces, and each is confined to a very narrow region consistent with how these data were generated. The two stability datasets become separated in the embedding space. 
The Tsuboyama stability dataset is divided into $\sim$64 individual sets, corresponding to the 64 DMSs in the dataset. Rocklin artificial proteins occupy their own region of space, and for the corresponding DTs, the best-performing layers also lie shallow in the model (\Cref{fig:perf}J). The third stability-related dataset, the Meltome Atlas dataset, occupies the same region as the majority of the data.

We chose the SCOPe40 dataset as a representative set of protein sequences that evenly samples the space of naturally occurring structured proteins and calculated the separation from it using pairwise silhouette scores of all other datasets (\Cref{fig:latent_sp}G). The two best separated datasets across all layers are the two large DMS datasets, fluorescence and GB1. The Rocklin stability dataset is also well separated, as it contains many artificial proteins and several DMS sets. In contrast, the Tsuboyama stability set is not separated from SCOPe40, but this is rather a deficiency of the metric used, since we observed in \Cref{fig:latent_sp}D-F that across the layers it falls apart into many small sets corresponding to individual proteins' DMSs. 
Here, we additionally considered two datasets of artificially designed proteins and their matching natural counterparts with experimentally measured activities: ProGen-generated and natural lysozymes~\cite{madani2023large} and bmDCA-generated and natural chorismate mutases~\cite{figliuzzi2018pairwise, russ2020evolution}. For both lysozyme and chorismate mutase datasets, we observe no separation in the shallow layers, but distinct separation in the deeper half of the PLMs, independent of enzyme provenance. 

Finally, we considered several datasets of proteins not present in the pre-training data (Supplementary Table 8) to test whether PLMs simply memorize these data or generalize from them, since most PLMs are trained on natural sequences from different redundancy-removed subsets of UniProt~\cite{bateman2024uniprot}. First, we used metagenomic protein sequences with less than 30\% sequence identity to UniProt from NovelMetaG~\cite{prabakaran2025deciphering}. We also assumed the artificial proteins were out-of-distribution, since they were not present in the training data due to their provenance. Additionally, we again analyzed the Rocklin stability dataset~\cite{rocklin2017global} that contains a large proportion of artificial sequences generated with Rosetta~\cite{rohl2004protein}.

For the NovelMetaG dataset, we compared losses for smaller and larger models with the SCOPe40 dataset, since no phenotypic data are available for NovelMetaG. In all cases, losses are similar in the shallow layers for both datasets, while in the deeper layers the loss for the SCOPe40 sequences drops much more rapidly (\Cref{fig:explanations}A and B).

Both artificial enzyme datasets have $\sim$70\% sequence identity to their closest UniProt neighbor. Yet across all layers, embeddings are not useful for predicting activity for artificial lysozymes (\Cref{fig:explanations}C), while they can be to some extent used to predict activity of both artificial and naturally occurring chorismate mutases relatively well (\Cref{fig:explanations}D). One may notice that in the shallow layers the chorismate mutase sequences are better mixed with the general protein population (\Cref{fig:latent_sp}G). The difference between the two sets of artificial proteins may lie in the generation process: the autoregressive ProGen~\cite{madani2023large} model for lysozymes and the direct coupling-based model bmDCA~\cite{figliuzzi2018pairwise} for chorismate mutases. Alternatively, the chorismate mutase dataset is more than 10 times larger, providing more training data.

For the Rocklin stability data, which comprises three parts (Rosetta-generated miniproteins, DMSs of 14 of these generated peptides, and DMSs of 3 natural short proteins), we trained separate probes in two ways: first, training one probe for the whole dataset and evaluating it on the subsets (\Cref{fig:explanations}E); and second, training a separate probe for each subset (\Cref{fig:explanations}F). For comparison, we also consider the Tsuboyama stability dataset that comprises DMSs for 64 natural proteins. In both cases, probe performance on artificial proteins without DMS is lower than on DMS-containing datasets (except for the probe trained and evaluated on the Rocklin subset with three DMSs of natural proteins, which contains only $\sim$2,000 data points). Notably, probes trained on the DMS subset for artificial proteins reach performance comparable to the Tsuboyama data. Hence, we believe it is more important that the probe model sees enough similar sequences during downstream training than that the pre-training dataset contains similar sequences.

\section{Discussion}

In this work, we present a comprehensive study of the informativeness of PLM embeddings across layers for various architectures and DTs, providing insights into where the information is stored. We show that the last layer is rarely the most useful for a DT, which runs counter to common practice. Taking the last layer is inherited from end-to-end networks, where every layer is optimized for the final objective and depth and performance increase together, as shown for image classifiers~\cite{alain2016understanding}. PLMs are trained differently: they are pre-trained on a self-supervised objective and only later connected to a downstream head, so the last layer need not be the most informative for a task the model was not trained on. Indeed, it has been reported earlier~\cite{kumar2025layer} that intermediate representations can outperform the last for a particular task, but without a systematic account across models and tasks. Here, we provide this account. 

The behavior we observe has a counterpart in natural language processing. It has been demonstrated that under a masked-language modeling objective, the information identifying the input token is first lost in the middle of a transformer-based network and then recovered in the top layers, a two-stage process called context encoding followed by token reconstruction~\cite{voita2019bottom}. The most informative representations therefore lie in the intermediate layers, while the deepest layers again reflect the pre-training objective. Importantly, we show that the different informativeness of PLM layers is largely insensitive to the data split, objective, difficulty, and probing model.

DTs are best served by different layers, and language models show a parallel: the stages of linguistic analysis are recovered in a fixed order along the network depth~\cite{tenney2019bert}. For PLMs, Vig et al. reported that shallow layers reflect simpler biophysical and secondary-structure features while deeper layers capture contacts and binding sites~\cite{vig2020bertology}. Our data are organized differently. Following previous research~\cite{kumar2025layer, vig2020bertology}, we show that alignment of the pre-training and DT objectives is crucial for the relevance of information extracted from shallow to deeper layers: the more they align, the more informative the deep layers are. We do not observe a fixed progression from simple to complex properties; instead, the depth of the best layer tracks how closely a task resembles the residue-level pre-training objective. In residue-level DTs, which align with the objective, embeddings improve with depth; in whole-protein and deep-mutational-scan (DMS) DTs, where objectives do not align, the most informative layers lie shallower. In general, performance across layers can serve as a proxy for how well the pre-training and DT objectives align. The same reasoning explains why fine-tuning improves performance so consistently~\cite{schmirler2024fine}: in addition to adapting the weights, it moves the task-relevant representation to the end of the network, where the prediction head reads it.

These results also have practical consequences for how we compare and use PLMs. Because the best layer differs between models, comparisons that evaluate every PLM at its last layer can rank them partly by how close the embeddings of the last layer happen to be to its own best layer, rather than by the information the model contains. A fairer comparison should consider the best layer for each PLM and be task-specific. Implementing this is inexpensive, since even a small fraction of a downstream dataset gives a good estimate of the informativeness across layers. This prompted us to create \verb+PLMSommelier+, a tool that samples a fraction of the data, estimates the most informative layer, and returns a truncated model with all embeddings computed down to that layer. In every case we examined, the truncated model was faster and smaller than the full one, with often a marginal increase in performance as well (\Cref{fig:benchmark_fig}). To make sure that \verb+PLMSommelier+ does not mislead the users in cases with unclear results, the tool reports its confidence levels, obtained through seed agreement and notifies the user if the accuracy of the predicted best layer can't be confirmed with high enough confidence. This makes sure that the tool can be used for virtually all PLM applications. 

The finding that DMS DTs are best predicted using embeddings from shallow layers may seem at odds with how PLMs are used in zero-shot variant-effect prediction, where the model's output is used to predict its masked-token probabilities~\cite{meier2021zeroshot}. The two are not in conflict. Zero-shot scoring uses the pre-training objective itself, whereas a supervised probe may rely on local sequence context already present in the early layers and not benefit from the deep, objective-specialized representations. If so, the appropriate layer depends not only on the task but on whether it is posed as a supervised probe or as a likelihood under the model.

We note that the best layer cannot be chosen based on factors such as the type of DT, intrinsic metrics of the latent space spanned by embeddings, or the general topology of the embedding space. The geometry of latent spaces has been characterized before~\cite{valeriani2023geometry}, and our results are consistent with this work: we use the same descriptors, but find that, on their own, they do not predict which layer is most useful for a task. The geometric descriptors summarize how a representation is organized, whereas a layer's usefulness depends on whether a simple downstream model can extract the relevant variation from it, and the two need not coincide. In our data, they do not, so we must identify the best layer empirically rather than predict it from the geometry.

We also asked whether layer behavior reflects how close the input proteins are to the pre-training distribution, i.e., whether the models memorize their training data or generalize from it. Our evidence does not support a single explanation. For metagenomic sequences with little similarity to UniProt~\cite{prabakaran2026quantifying} and for the Rosetta-designed proteins of the Rocklin set~\cite{rocklin2017global}, which are out-of-distribution for the pre-training data of most PLMs, the loss does not drop, or embeddings perform poorly, particularly in the deeper layers. For the two sets of designed enzymes~\cite{madani2023large, russ2020evolution}, the result differs: the activity of the artificial lysozymes is predicted weakly at every layer, and that of the artificial chorismate mutases is about as good as for their natural counterparts. The enzyme sets are small, so the weak lysozyme signal may reflect the limited and noisy phenotype data rather than the representations. Consistent with the absence of a clear effect, natural and designed enzymes in the same family separate from the bulk of natural proteins to the same degree in the deeper layers, so this separation reflects how narrow or unusual a dataset is rather than whether its members are designed. Hence, we do not consider proximity to the pre-training distribution to be the factor governing layer behavior. One observation that holds for these analyses concerns the downstream data: predictability depends on whether the downstream training set contains enough similar sequences, not on whether the pre-training set did.

We cannot determine from these data why the designed sets differ among themselves. One possibility, which we cannot test here and which would require larger phenotype-labeled design sets, is whether it matters how well the designed sequences match the natural sequence statistics. Of the three sets that we used, lysozymes were created with ProGen, a PLM~\cite{madani2023large}, chorismate mutases using a direct-coupling statistical model~\cite{russ2020evolution}, and Rocklin miniproteins with the physics-based Rosetta model~\cite{rohl2004protein}. This is consistent with Prabakaran and Bromberg's report that computationally designed proteins can appear biologically plausible to these models~\cite{prabakaran2026quantifying}. Our analysis also goes beyond theirs: they evaluated embeddings at a single, final layer against a randomized-sequence baseline, whereas we resolve informativeness layer by layer, against a reference of natural proteins, and on functional rather than shuffled sequences. This has a practical implication for using these models to score or filter designed proteins: a model that represents a natural-like design as it does a natural protein cannot distinguish the two.

Two caveats apply to our work. Probing measures what a simple model can extract from a representation, not everything the representation contains, and a more expressive probe can recover more~\cite{hewitt2019designing}. We used simple probes and confirmed that our conclusions hold across probe types, task difficulty, and data splits, but they remain statements about decodable information. Our conclusions about provenance also rest on a small number of designed proteins with measured activities, and the autoregressive models, which behave differently from the masked ones, are represented by fewer examples. Within these limits, our study systematically accounts for where PLM task-relevant information is most accessible and why the depth of the corresponding layer varies, relates this to what is known about masked language models more generally, and provides a tool that uses it. Why one protein-level task is best served by a particular layer rather than another remains open, and is the question we consider most worth pursuing next.

\section{Methods}

\subsection{Mathematical foundations}
Protein language models (PLMs) follow mostly the same principles as large language models in natural language processing. Input is a sequence of amino acids $\left(a_i\right)_{i=1}^m$. For most PLMs, each amino acid corresponds to a token, $t_j=a_i$, but ProtGPT2 utilizes a byte-pair encoding (BPE) tokenizer~\cite{gage1994new} grouping multiple amino acids into one token, $t_j=(a_k \dots a_l)$. The resulting sequence of tokens $\left(t_j\right)_{j=0}^n$ is then processed by an $L$-layered PLM encoder yielding the token-wise layer activations $\ve{h}_{i,j}^l$, representing token $t_j$ in protein $i$ of layer $l$\footnote{ProtGPT2 cannot be applied to amino acid-level prediction tasks because its token embeddings represent multiple amino acids due to BPE tokenization. Therefore, ProtGPT2 only produces whole-protein embeddings.}. For the remainder of this section, we assume a token corresponds to one amino acid and use the terms interchangeably in this context. For whole-protein prediction tasks, we use mean pooling to compute per-layer embeddings $\ve{h}_i^l$ of a protein $i$ as:

\begin{align*}
    \ve{h}_i^l=\frac{1}{n}\sum\limits_{j=0}^n\ve{h}_{i,j}^l.  
\end{align*}

We call the vector space $\vs{H}^l$ over all protein embeddings $\ve{h}^l_i$ the \emph{layer-latent space} for a dataset and the specific layer $l$ of a PLM.

\subsubsection{Latent space metrics}\label{sec:metrics}
We use three metrics to analyze the complexity and structure of $\vs{H}^l$ and the changes between $\vs{H}^l$ and $\vs{H}^{l+1}$. The intrinsic dimension (ID) is not defined in a standard way, but the general idea is to estimate the minimum number of dimensions needed to represent the full complexity of a latent space. In our work, we follow the lead of Valeriani et al.~\cite{valeriani2023geometry} and use the TwoNN estimator~\cite{facco2017estimating} to compute the IDs. For each $\ve{h}_i^l$, it requires the distances $r_{i,1}$ and $r_{i,2}$ to the two closest neighbors in $\vs{H}^l$. The ratio $\mu_i=\sfrac{r_{i,2}}{r_{i,1}}$ thereof follows a Pareto distribution $P(x,\alpha)$ whose shape parameter $\alpha$ is equal to the ID.

The second measure of latent space complexity is \emph{variance@10}, which denotes the portion of variance explained by the first 10 principal components in a principal component analysis.

Third, the Neighborhood Overlap (NO) $\chi_k^{l,m}$ measures how much the $k$-neighborhoods change between two latent spaces $\vs{H}^l$ and $\vs{H}^m$~\cite{doimo2020hierarchical}. Let $\mathcalorig{N}_{k,i}^l$ be the set of the nearest $k$ neighbors of $i$ in $\vs{H}^l$, then $\chi_k^{l,m}$ is defined as

\begin{align*}
    \chi_k^{l,m}=\frac{1}{N\cdot k}\sum\limits_{i=1}^N\left|\mathcalorig{N}_{k,i}^l\cap\mathcalorig{N}_{k,i}^m\right|.
\end{align*}

The higher the values for $\chi_k^{l,l+1}$, the less the compared latent spaces change. Intuitively and empirically demonstrated in this study and \cite{valeriani2023geometry}, the layers of PLMs produce similar latent spaces (high $\chi_k^{l,l+1}$) because each layer adds another layer of abstraction on top of the previous knowledge \cite{zeiler2014visualizing} rather than reorganizing the latent space. Following Valeriani et al., we use $k=10$ in our experiments and find that $\chi_k^l$ is stable across different values of $k$~\cite{valeriani2023geometry}.

\subsubsection{Layer probes}
Following Alain \& Bengio \cite{alain2016understanding}, we define linear \emph{layer probes} with weights $\ve{w}$ and bias $b$ as:

\begin{align*}
    f^l:\vs{H}^l&\rightarrow\mathbb{L}\quad\text{with}\\
    \ve{h}_i^l&\mapsto \texttt{act}(\ve{w}\cdot\ve{h}_i^l + b),
\end{align*}

where $\mathbb{L}$ is the label space of a dataset: $\mathbb{R}$ for regression tasks and $[0,1]^D$ for $D$-dimensional classification problems, \texttt{act} is an activation function, the identity function for regression tasks, a softmax function for binary and multi-target classification problems, and a class-wise sigmoid function for multi-label problems. For amino acid-level tasks, $f_l$ maps from the space of amino acid embeddings $\vs{h}^l$ and is defined analogously. Since the latent spaces are not necessarily linearly separable, we also use $k$-nearest neighbor probes operating on $\vs{H}^l$ with $k=10$.

\subsubsection{Sparse layer probes}
To estimate how little data is necessary to identify an almost-best layer, we trained sparse layer probes $f^l_\text{sp}$ on sparse vector spaces $\vs{H}^l_\text{sp}$ for all layers $l\le L$ of a PLM. From that, we identify the $f^{l\ast}_\text{sp}$ with the best performance on a DT. Then, we compute the ratio of the performance of $f^l$ and $f^{l\ast}$, both trained on $\vs{H}^l$. This yields the relative-to-best performance metric used in \Cref{fig:ablation}E-H.

\subsection{PLMSommelier}

The \verb+PLMSommelier+ Python package works as both a Python package and a command-line interface.
It uses sparse-layer probe logic, taking a small fraction of the dataset (15\% as a default setting) and running a small probe on embeddings from each layer. 
It then selects the best layer for a given downstream task. 
To account for noise, if multiple layers are within 2\% performance, we select the shallowest layer, following Occam's Razor, which also saves compute.
This procedure is repeated with 5 random seeds, and the tool returns a truncated model that is not only better than the last-layer version but also faster.
If the seeds show a strong disagreements, the user is notified and an increase in the sampling size is suggested.
However, for the datasets described in this paper all application of \verb+PLMSommelier+ yielded either moderate or high confidence.
The tool works with all PLMs available on Huggingface and can be easily extended to include support for any in-house PLM.
It is available at \href{https://github.com/kalininalab/PLMSommelier}{GitHub} and additionally can be installed via \textit{pip}.

\subsection{Protein Language Models}\label{sec:plm}
\begin{table*}[t!]
    \centering
    \caption{\textbf{Comparison of investigated PLMs.} \\ {\normalfont $^\text{\textdagger}$ ProstT5 is the ProtT5 model fine-tuned on translating the amino-acid sequences to 3Di sequences and back for 17M structures from the AlphaFold database.}}
    \label{tab:plms}
    \resizebox{\textwidth}{!}{%
    \begin{tabular}{llllrrr}
        \toprule
        \textbf{PLM} & \textbf{NLP Arch} & \textbf{Pre-training Datasets} & \textbf{seen data} & \textbf{\# layer} & \textbf{\# weights [M]} & \textbf{Embed. Dim.} \\ \midrule
        \multicolumn{7}{l}{\textit{masked-language model loss (bi-directional)}} \\
        ESM-2~8M & BERT & UniRef50 \& UniRef90 & 1T AAs & 6 & 12 & 320 \\
        ESM-2~35M & BERT & UniRef50 \& UniRef90 & 1T AAs & 12 & 35 & 480 \\
        ESM-2~150M & BERT & UniRef50 \& UniRef90 & 1T AAs & 30 & 150 & 640 \\
        ESM-2~650M & BERT & UniRef50 \& UniRef90 & 1T AAs & 33 & 650 & 1280 \\
        ESM-2~3B & BERT & UniRef50 \& UniRef90 & 1T AAs & 36 & 3,000 & 2560 \\
        
        ESMC-300m & Transformer & UniRef70, MGnify 70, JGI 70 & 6.2T AAs & 30 & 300 & 960 \\
        ESMC-600m & Transformer & UniRef70, MGnify 70, JGI 70 & 6.2T AAs & 36 & 600 & 1152 \\
        ProtT5 & T5 & BFD 100, then UniRef50 & 4.9B + 2B seqs & 24 & 3,000 & 1024 \\ 
        ProstT5 & T5 & 17M non-redundant AFDB structures$^\text{\textdagger}$ & 102M + 16.8M seqs & 24 & 3,000 & 1024 \\ \midrule
        \multicolumn{7}{l}{\textit{next-token prediction language loss (auto-regressive, generative)}}\\
        ProGen2-small & Transformer & UniRef90 \& BFD30 & 175B seqs & 12 & 151 & 1024 \\
        ProGen2-medium & Transformer & UniRef90 \& BFD30 & 175B seqs & 27 & 764 & 1536 \\
        ProGen2-large & Transformer & UniRef90 \& BFD30 & 200B seqs & 32 & 2,700 & 2560 \\
        ProtGPT2 & GPT & UniRef50 & 2.5B seqs & 36 & 738 & 1280 \\
        \bottomrule
    \end{tabular}%
    }
\end{table*}

In this work, we investigate protein language models (PLMs) across five families, spanning a wide range of architectures, training paradigms, model sizes, and embedding dimensions. The ESM-2~\cite{lin2023evolutionary}, ProtT5\cite{elnaggar2021prottrans}, and ProstT5~\cite{heinzinger2024bilingual} models are trained with the \emph{masked language model (MLM) loss}, while ProGen2~\cite{nijkamp2023progen2} and ProtGPT2~\cite{ferruz2022protgpt2} were trained with the generative \emph{next-token prediction (NTP) loss}. Due to experimental feasibility and hardware constraints in storing embeddings, we focus only on models with a maximum parameter count of 3 billion. An overview of the technical details of the PLMs is given in \Cref{tab:plms}.

\subsubsection{ESM} The models from the Evolutionary Scale Modeling (ESM) initiative by Facebook AI, later independently as EvolutionaryScale, were the first PLMs to consistently achieve state-of-the-art performance across various tasks~\cite{rives2021biological}. These models are based on the encoder-only BERT architecture~\cite{devlin2019bert}. In our work, we use the 8M, 35M, 150M, 650M, and 3B models from the ESM-2 selection, as well as ESMC-300M and ESMC-600M~\cite{esm2024esmcambrian}. The ESM-2 models were trained on 1 trillion tokens by first sampling proteins from UniRef50~\cite{bateman2024uniprot}, then, for each sequence, from the corresponding UniRef90 clusters. The ESMC models were trained on 6.2 trillion tokens sampled from a combination of the UniRef, MGnify~\cite{richardson2023mgnify}, and JGI databases~\cite{nordberg2014jgi}, each clustered at 70\% sequence identity.

\subsubsection{ProtT5} In the ProtTrans paper, the RostLab tested six different LLM architectures for their transferability to the ``protein language''. The pre-training was performed on different combinations of the Big Fantastic Database (BFD100)~\cite{steinegger2019bfd} and UniRef (UniRef100 and UniRef50), clustered at 50\% or 100\% pairwise sequence identity. The T5-based ProtT5-XL and ProtT5-XXL models emerged as the best. In this work, we use the ProtT5-XL-U50 and will refer to it as the ProtT5 model. This model was first pre-trained on 4.9 billion sequences from the BFD and then fine-tuned on 2 billion sequences from UniRef50.

\subsubsection{ProstT5} The RostLab published another PLM based on their findings in the ProtTrans paper. For ProstT5, they built a dataset from AlphaFoldDB containing pairs of amino acid sequences and 3Di sequences of one protein. The 3Di tokens, computed with FoldSeek, represent the protein structure as a sequence. The authors then fine-tune the ProtT5 model first on the unmasking of both AA and 3Di sequences, and then on the bidirectional translation task between the two formats.

\subsubsection{ProGen2} Multiple works show that latent spaces are structured differently for models trained bidirectionally or auto-regressively~\cite{valeriani2023geometry, vig2020bertology, skean2025layer, saponati2025underlying}. To compare the effect of different pre-training objectives, we also include four generative PLMs. The ProGen2 models were trained on the combination of UniRef90 and BFD30 with the next-token prediction loss. Here, we compare only the small, medium, and large models, which have been trained on 175-200 billion sequences. These numbers are much higher than for the bi-directional models because they include many shorter sequences to predict early amino acids.

\subsubsection{ProtGPT2}\label{sec:protgpt2} An important first step in NLP is to tokenize the input sequences to make it easier to pick up common motifs in the text. For PLMs, subword tokenization is often omitted to generate embeddings for each amino acid. ProtGPT2 is one of the few PLMs that use byte-pair encoding for subword tokenization. BPE computed 50,256 tokens in the Swiss-Prot (version 2021\_04), which were then applied to the UniRef50 dataset before pre-training.

\begin{table*}[t]
    \centering
    \caption{\textbf{Summary of the benchmarked downstream datasets.} For each dataset, we provide structural levels, sizes, and downstream tasks, as well as the mean sequence identity measured as pairwise sequence similarity using MMseqs2. The clusters were computed with CD-HIT at a maximum 40\% sequence similarity~\cite{li2006cd}.}
    \label{tab:datasets_summary}
    \resizebox{\textwidth}{!}{%
    \begin{tabular}{llccp{5cm}cc} \toprule
        \textbf{Dataset}    & \textbf{Level} & \textbf{\#Seqs} & \textbf{Mean Length} & \textbf{Associated Downstream Tasks}          & \textbf{Mean Seq. Identity (\%)} & \textbf{\# clusters} \\ \midrule
        Fluorescence        & Protein        &  54,024         & 237                  &  1. Intensity Regression                      & 0.9671 & 1 \\
                            &                &                 &                      &  2. Active/Dead Classification                &  \\ \addlinespace
        GB1                 & Protein        & 149,361         & 56                   &  3. Binding Fitness Regression                & 0.8841 & 1 \\ \addlinespace
        Rocklin stability   & Protein        &  68,976         & 45                   &  4. Stability Score Regression                & 0.2388 & 11,342 \\ \addlinespace
        Tsuboyama stability & Protein        & 117,811         & 59                   &  5. Stability Regression                      & 0.2192 & 63 \\ \addlinespace
        DeepSol             & Protein        &  71,094         & 295                  &  6. Binary Solubility Classification          & 0.1653 & 44,523 \\ \addlinespace
        SCOPe40 2.08        & Protein        &  15,176         & 185                  &  7. Fold Classification                       & 0.1587 & 13,901 \\
                            &                &                 &                      &  8. Superfamily Classification                &  \\
                            & Residue        &                 &                      &  9. DSSP 3-Class Prediction                   &  \\
                            &                &                 &                      & 10. DSSP 8-Class Prediction                   &  \\ \addlinespace
        Ligand Binding      & Residue        &  12,317         & 454                  & 11. Residue-Level Interaction Classification  & 0.1404 & 7,540 \\ \addlinespace
        Meltome Atlas       & Protein        &  30,231         & 475                  & 12. Species Classification                    & 0.1373 & 15,579 \\ 
                            &                &                 &                      & 13. Melting Temperature ($T_\text{m}$) Regression    &  \\ \addlinespace
        DeepLoc2.0          & Protein        &  28,302         & 557                  & 14. Binary Membrane Protein Classification    & 0.1366 & 17,645 \\
                            &                &                 &                      & 15. 10-Class Subcellular Localization         &  \\ \bottomrule
    \end{tabular}%
    }
\end{table*}

\subsection{Datasets}\label{sec:datasets}

Fluorescence~\cite{sarkisyan2016local}, Rocklin stability~\cite{rocklin2017global} and SCOPe40-derived~\cite{chandonia2022scope} datasets follow TAPE~\cite{rao2019evaluating}; GB1~\cite{olson2014comprehensive}, Tsuboyama stability~\cite{tsuboyama2023mega}, DeepSol~\cite{khurana2018deepsol}, Meltome Atlas~\cite{jarzab2020meltome}, DeepLoc2.0~\cite{thumuluri2022deeploc} and the ligand-binding dataset~\cite{littmann2021protein} were added from their original papers.
An overview of key information on the datasets is given in \Cref{tab:datasets_summary}.
We downloaded the Fluorescence, Rocklin stability, DeepSol, Meltome Atlas, and DeepLoc2.0 datasets from the HuggingFace website. The exact vendors are listed in the data availability section.

\begin{figure}[t]
    \centering
    \includegraphics[width=\linewidth]{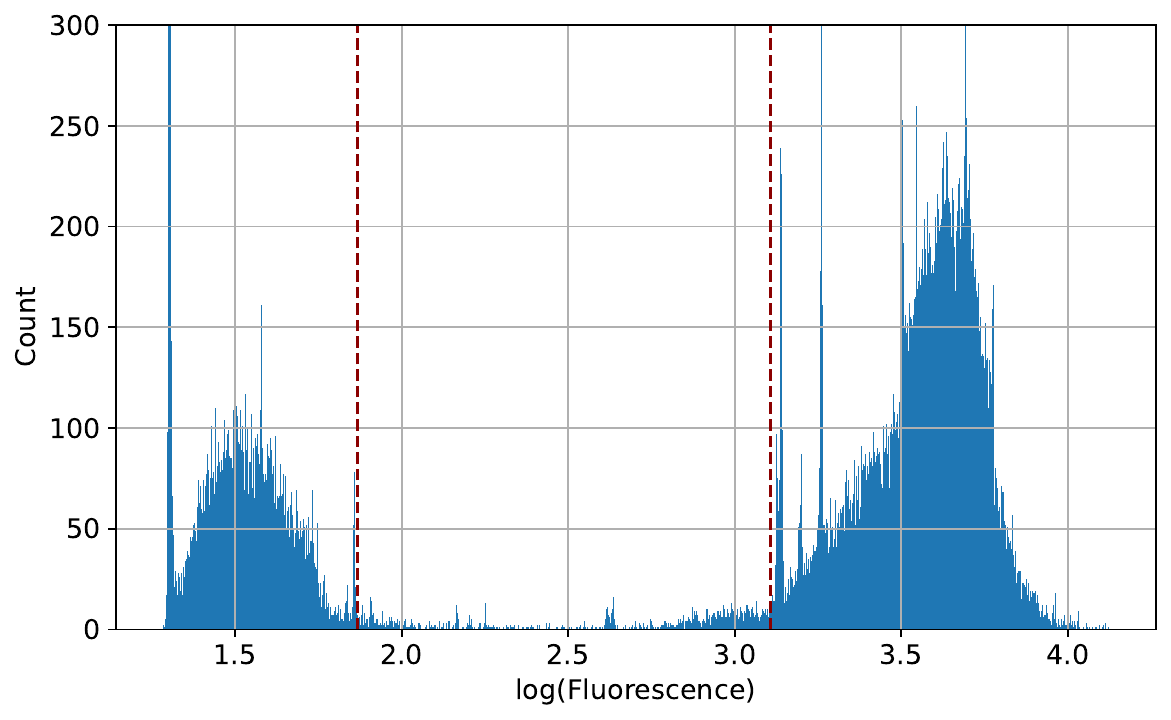}
    \caption{\textbf{Label distribution in the Fluorescence dataset.} The blue histogram shows the distribution of log-fluorescence values. The dashed red lines highlight the decision boundaries for the conversion from regression to classification. The data falling in between have been omitted.}
    \label{fig:fluorescence}
\end{figure}

\subsubsection{Fluorescence}

Derived from the systematic random mutagenesis of the \textit{Aequorea victoria} green fluorescent protein (GFP) by Sarkisyan et al.~\cite{sarkisyan2016local}, this dataset evaluates how amino acid substitutions affect a protein's biochemical phenotype. It includes approximately 54,000 mutant sequences, each paired with its experimentally measured log-fluorescence intensity. Originally, the task was designed as a regression problem, but the distribution of the log-fluorescence values is bimodal, representing the active and inactive mutants. For that reason, we use this dataset for both classification and regression tasks. The conversion to a classification dataset was performed by fitting a Gaussian Mixture Model to the data, computing 2-standard-deviation intervals around the means, and separating into the upper bound of the lower mode and the lower bound of the upper mode (\Cref{fig:fluorescence}). The data between these two bounds is omitted because, when converting a regression problem to classification, the task is easier to learn when the two classes are separated by an interval rather than by a single threshold.

\subsubsection{GB1}
This dataset resulted from a deep mutational scan investigating the epistatic effects of mutations at 4 positions in the IgG-binding domain of protein G. It includes $\sim$149,000 mutants and measures the fold change of each mutant relative to the wild-type protein. Similar to the Fluorescence regression task, we predict mutation effects~\cite{olson2014comprehensive}.

\subsubsection{Rocklin stability}\label{sec:stability}
Sourced from the high-throughput screening benchmark developed by Rocklin et al.~\cite{rocklin2017global} and curated within the TAPE framework~\cite{rao2019evaluating}, this dataset evaluates the thermodynamic stability of over 69,000 unique short miniproteins (40–50 amino acids in length). The sequence space comprises a diverse mixture of computationally generated \textit{de novo}-designed miniproteins, stable natural control domains (including the Pin1 WW-domain, hYAP65 WW-domain, villin headpiece, and BBL protein), systematically introduced single-residue point mutants, and unfolded negative controls. The associated global regression task requires models to predict a continuous ``Stability Score'' derived from multiplexed protease susceptibility assays. This metric quantifies how effectively a surface-displayed protein resists enzymatic degradation by trypsin and chymotrypsin, serving as a direct proxy for folding energetics and structural robustness.

\subsubsection{Tsuboyama stability}
For this dataset, Tsuboyama et al. measured the thermodynamic folding stability of $\sim$900,000 protein domains (average length: 59 amino acids). The resulting 1.8 million measurements were curated into a set of $\sim$776,000 high-quality folding stabilities, measured as $\Delta\Delta G$, in the original publication, covering all single-mutation variances and some double mutants~\cite{tsuboyama2023mega}. In this work, we use the further curated dataset from ProteinGym comprising DMS scans of 117,811 amino acid sequences from 64 proteins (635 to 5586 mutants per protein)~\cite{notin2023proteingym}.

\subsubsection{Meltome Atlas}
Similar to the dataset assembled by Rocklin et al. and Tsuboyama et al., the Meltome Atlas measures the thermostability of various proteins~\cite{jarzab2020meltome}. The main difference lies in the source of proteins: while the dataset compiled by Rocklin et al. consists predominantly of \textit{de novo}-designed short proteins (under 50 amino acids in length), the Meltome Atlas comprises 30,321 natural proteins from 13 different species. The labels are the melting temperatures of proteins in degrees Celsius, and the species are specifically chosen to cover the full range of preferred temperatures, including thermophilic bacteria.

\subsubsection{SCOPe40 2.08}

This dataset originates from the SCOPe40 dataset, the SCOPe database~\cite{chandonia2022scope} clustered at 40\% sequence similarity. We further filtered it to include only folds or superfamilies with more than 10 samples, removing sparse classes, resulting in $\sim$13,000 experimentally determined protein sequences. The protein-level tasks are the prediction of protein fold and superfamily from the SCOPe40 classification. Additionally, we ran DSSP~\cite{kabsch1983dictionary} on the proteins in this dataset, generating 3- and 8-class secondary-structure assignments, which are predicted at the residue level.

\subsubsection{DeepLoc2.0}

The DeepLoc2.0 dataset consists of $\sim$28,000 redundancy-reduced, eukaryotic proteins with experimentally verified subcellular localization tags \cite{thumuluri2022deeploc}. These comprise a binary classification task distinguishing between soluble and membrane-bound structural forms, and a 10-class multi-class classification task mapping whole proteins to their definitive cellular compartments (such as the nucleus, cytoplasm, or mitochondrion).

\subsubsection{DeepSol}
Lastly, DeepSol consists of $\sim$71,000 binary protein solubility labels~\cite{khurana2018deepsol}. Khurana et al. preprocessed the dataset to filter out homologous sequences and to remove unwanted bias between the independent test set and the training set.

\subsection{Technical requirements}
All experiments were conducted on 2 NVIDIA RTX3090 GPUs with 24 GB of RAM, 2 NVIDIA Tesla V100 GPUs with 16 GB of RAM, and the Saarland Informatics Campus High Performance Computing Cluster with multiple NVIDIA A100 GPUs with 40 GB of RAM. All of them used the CUDA v12.9 GPU driver. In total, we consumed $\sim$10 TB of storage space for the protein embeddings and 9.8 TB for the residue embeddings. We did not run the residue-level experiments for the ESM-2 3B, ProGen2-large, and ProtGPT models, but we estimate their memory consumption at an additional 10 TB. On the software side, we relied on PyTorch v2.9.1 and CuML v25.12.0. A full list of dependencies is provided in the requirements.txt file of the GitHub repository linked below.

For fine-tuning ESM-2 150M, we fine-tuned the full model using the AdamW optimizer [ref] with a learning rate of 0.0001, a batch size of 32, and a maximum of 100 epochs, with early stopping after 10 epochs without improvement in validation loss. We selected the final model with the best validation loss.

\section*{Data Availability}
All code to reproduce the findings of this project is available at \href{https://github.com/kalininalab/MilleFeuille}{https://github.com/kalininalab/MilleFeuille}. Additionally, we uploaded all processed datasets to Zenodo \href{https://doi.org/10.5281/zenodo.21869125}{https://doi.org/10.5281/zenodo.21869125}~\cite{zenodo_data}.
The code for PLMSommelier is available at \href{https://github.com/kalininalab/PLMSommelier}{https://github.com/kalininalab/PLMSommelier}. The package is installable from PyPI with \verb+pip install plmsommelier+.

\noindent Some of the raw datasets were acquired from the following HuggingFace vendors:
\begin{itemize}[leftmargin=*]
    \item Fluorescence: \href{https://huggingface.co/datasets/proteinea/fluorescence}{https://huggingface.co/datasets/proteinea/fluorescence}
    \item GB1: \href{https://huggingface.co/datasets/SaProtHub/Dataset-GB1-fitness}{https://huggingface.co/datasets/SaProtHub/Dataset-GB1-fitness}
    \item Rocklin stability: \href{https://huggingface.co/datasets/proteinglm/stability_prediction}{https://huggingface.co/datasets/proteinglm/stability\_prediction}
    \item Meltome Atlas: \href{https://huggingface.co/datasets/cradle-bio/meltome_cluster_split}{https://huggingface.co/datasets/cradle-bio/meltome\_cluster\_split}
    \item DeepLoc2.0: \href{https://huggingface.co/datasets/bloyal/deeploc}{https://huggingface.co/datasets/bloyal/deeploc}
    \item DeepSol: \href{https://huggingface.co/datasets/proteinea/solubility}{https://huggingface.co/datasets/proteinea/solubility}
\end{itemize}

SCOPe40 was taken directly from the website; the Tsuboyama stability data were extracted from ProteinGym, and the binding, lysozyme, and chorismate mutase datasets were sourced directly from the respective manuscripts.

\section*{Declarations}
\textbf{Author Contributions:} R.J., I.S., and O.V.K. conceived the project. R.J. implemented most of the code and conducted most of the experiments. I.S. conducted the experiments on the fluorescence and stability datasets and implemented \verb+PLMSommelier+. A.K. fine-tuned the ESM-2 150M model. D.K. and O.V.K. jointly supervised the project. All authors wrote and revised the manuscript.\\
\textbf{Competing Interests:} The authors declare no competing interests.

\section*{Acknowledgments}
We thank Wim Vranken in particular for discussing the Meltome Atlas and for valuable hints and ideas. 
We thank Fyodor Kondrashov, R Prabakaran, and Yana Bromberg for feedback on the manuscript.
We thank Alper Yurtseven for testing the PLMSommelier package.
We first tested the idea for this project in summer school projects in 2023-2025. The students who contributed to the projects' early phase over the three years were Anastasia Lavova, Olga Balakina, Inesa Grigoryan, Igor Chekmarev, Arina Matveichuk, Maria Shchekanova, Veronica Sharonova, Alisa Soloveva, Daniel Korkin, Igor Larin, Semen Karaban, Mariia Sobko, and Evgeniia Samokhvalova. R.J., I.S., and A.K. supervised these projects, along with Vera Terenteva, Alper Yurtseven, and Guangyi Chen.

\bibliographystyle{unsrt}
\bibliography{pnas-sample}

\ifbuildpreprint
  \renewcommand{\figurename}{Supplementary Figure}
\renewcommand{\tablename}{Supplementary Table}
\setcounter{figure}{0}

\begin{figure*}[th!]
    \centering
    \includegraphics[width=\textwidth]{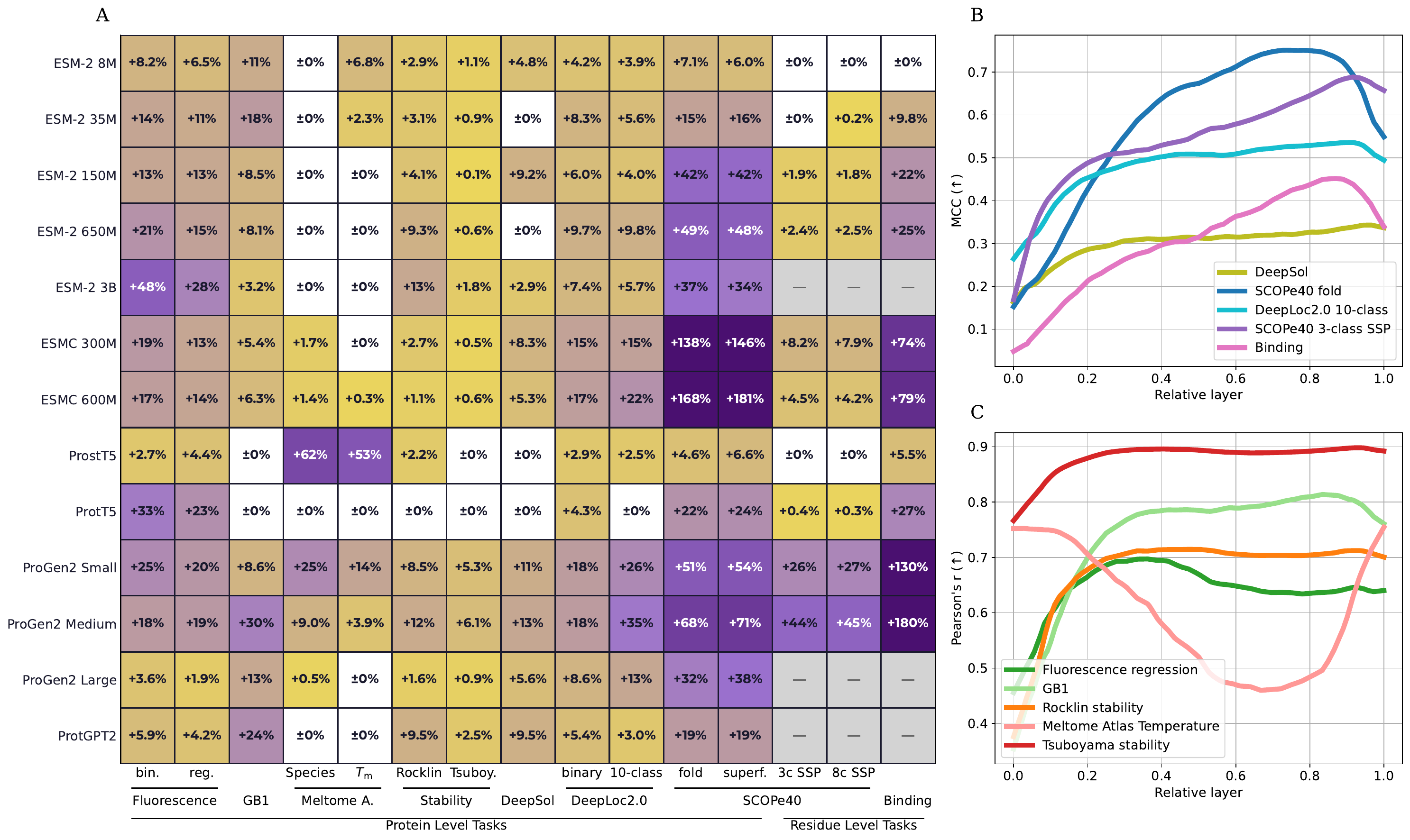}
    \caption{\textbf{Figure 1H-J with $k$-NN probes.} \textbf{A} Relative improvement of the best layer of each PLM over the last layer on all DTs with $k$-NN probes. ESM2-3B and ProGen2-Large required too much storage to compute residue-wise embeddings; ProtGPT2 has no per-residue embeddings, thus having no results for those tasks. $\pm0\%$ improvement implies the last layer was the best one. \textbf{B} and \textbf{C}: Average performance for each dataset across all models for classification (B) and regression (C) tasks. \\The overall trends are the same as in Figure 1, but Meltome Atlas marks an exception due to the extreme drop in performance after the first 10\% of model depth and is also the only dataset in which the first layer was not the worst predictor. Whereas the target variable here is essentially identical to the one in the Stability dataset, the behavior across PLMs' layers is very different. This can be explained by the fact that in the Meltome Atlas, the melting points of different proteins are strongly correlated with the species from which the protein comes, which in turn correlates with the amino acid composition~\cite{lopez2025supervised}. Naturally, the shallowest layers have the best access to amino acid composition, before the local context of each residue is taken into account. Furthermore, the last layer of both MLM and autoregressive models is more strongly connected to amino acid identity, since it is where the model must predict the amino acid type.\\Supplementary Figure 3 shows B and C in per-model resolution.}
\end{figure*}

\begin{figure*}[th!]
    \centering
    \includegraphics[width=\textwidth]{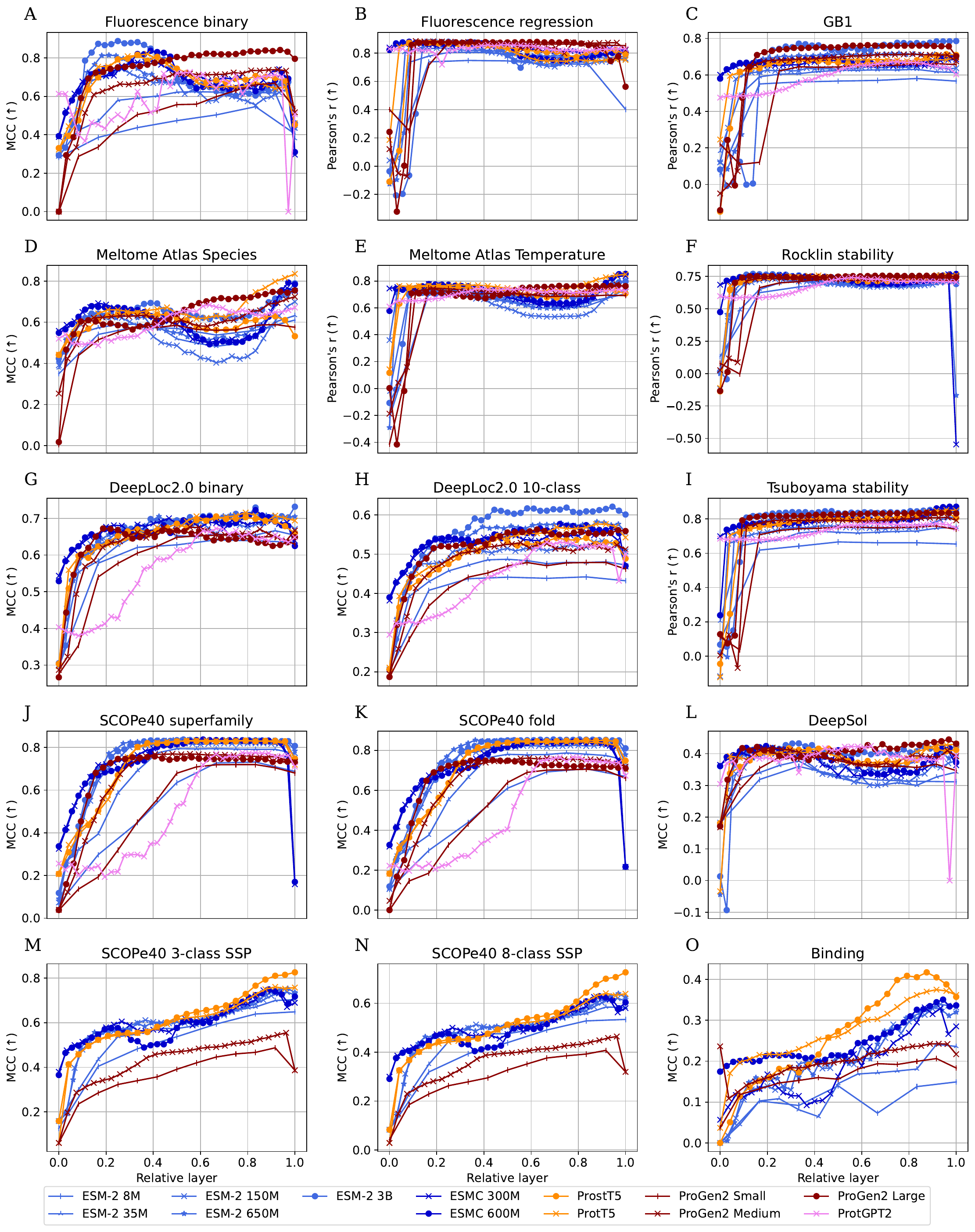}
    \caption{\textbf{Full layer performances.} Full model resolution of Figure 1 I and J with linear probes.}
\end{figure*}

\begin{figure*}[th!]
    \centering
    \includegraphics[width=\textwidth]{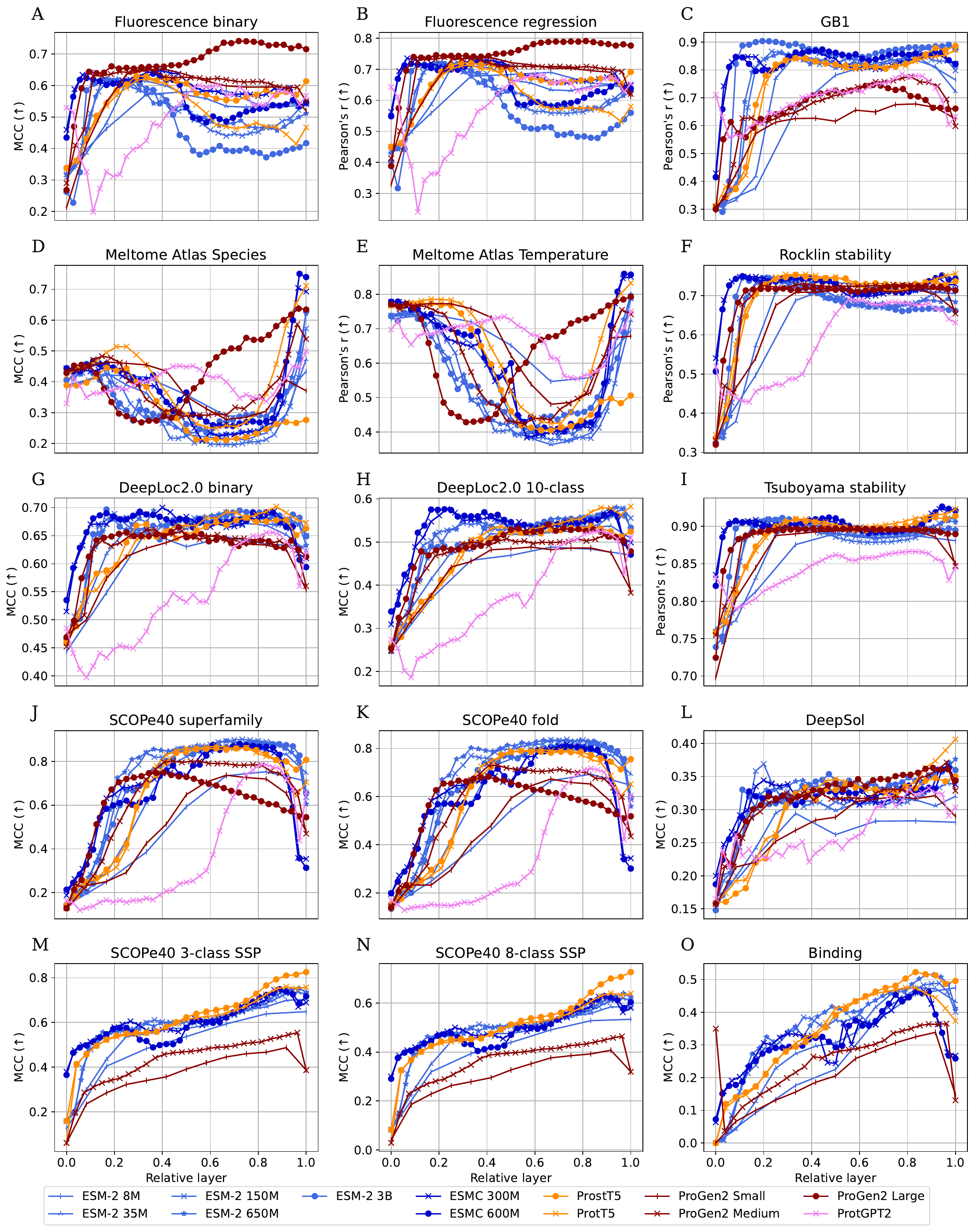}
    \caption{\textbf{Full layer performances.} Full model resolution of Supplementary Figure 1 B and C with $k$-NN probes.}
\end{figure*}

\begin{figure*}
    \centering
    \includegraphics[width=\textwidth]{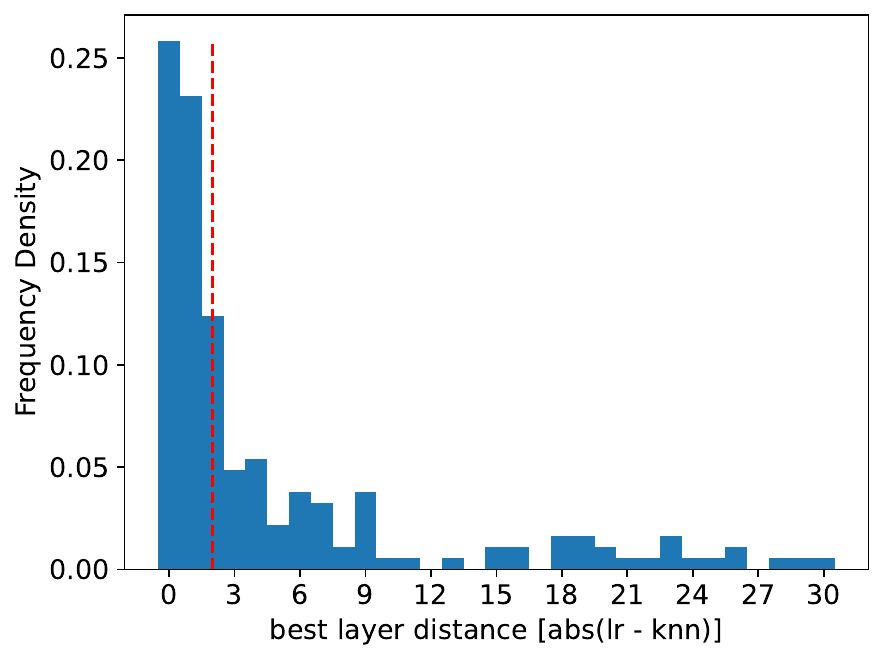}
    \caption{\textbf{Distance of best performing layers of LR and $k$-NN probes. The dashed line marks the median at 2.}}
\end{figure*}

\clearpage
\begin{table*}[th!]
    \centering
    \caption{Comparison of the performance of linear probes of the last layers of all PLMs and all tasks. Regression tasks (marked $^\dagger$) report Pearson correlation (PCC); all other tasks report Matthews correlation coefficient (MCC). ESM2-3B and ProGen2-Large required too much storage to compute residue-wise embeddings; ProtGPT2 has no per-residue embeddings, thus having no results for those tasks.}
    \resizebox{\textwidth}{!}{%
    \begin{tabular}{lcccccccccccc|ccc}
        & \multicolumn{12}{c|}{\textsc{protein-level tasks}} & \multicolumn{3}{c}{\textsc{residue-level tasks}}\\
        \textbf{Models} & \multicolumn{2}{c}{\textbf{Fluorescence}} & \textbf{GB1}$^\dagger$ & \multicolumn{2}{c}{\textbf{Meltome Atlas}} & \multicolumn{2}{c}{\textbf{Stability}} & \textbf{DeepSol} & \multicolumn{2}{c}{\textbf{DeepLoc2.0}} & \multicolumn{2}{c|}{\textbf{SCOPe40}} & \multicolumn{2}{c}{\textbf{SSP}} & \textbf{Binding} \\
        & Bin. & Reg.$^\dagger$ &  & $T_\text{m}^\dagger$ & Species & Rocklin$^\dagger$ & Tsubo.$^\dagger$ &  & Bin. & 10c & Fold & SF. & 3c & 8c & \\ \midrule
        ESM-2 8M & 0.405 & 0.402 & 0.566 & 0.611 & 0.704 & 0.706 & 0.654 & 0.312 & 0.635 & 0.432 & 0.669 & 0.687 & 0.635 & 0.516 & 0.149 \\
        ESM-2 35M & 0.380 & 0.778 & 0.613 & 0.632 & 0.739 & 0.724 & 0.752 & 0.342 & 0.654 & 0.472 & 0.751 & 0.771 & 0.696 & 0.574 & 0.236 \\
        ESM-2 150M & 0.436 & 0.761 & 0.637 & 0.699 & 0.769 & 0.734 & 0.813 & 0.379 & 0.671 & 0.507 & 0.734 & 0.750 & 0.743 & 0.631 & 0.331 \\
        ESM-2 650M & 0.435 & 0.809 & 0.706 & 0.761 & 0.795 & -0.168 & 0.845 & 0.391 & 0.705 & 0.556 & 0.778 & 0.783 & 0.776 & 0.671 & 0.320 \\
        ESM-2 3B & 0.460 & 0.795 & \textbf{0.786} & 0.789 & 0.798 & 0.691 & 0.859 & 0.423 & \textbf{0.732} & \textbf{0.601} & \textbf{0.810} & \textbf{0.807} & -- & -- & -- \\
        ESMC 300M & 0.296 & 0.822 & 0.685 & 0.752 & 0.837 & -0.547 & 0.857 & 0.364 & 0.639 & 0.469 & 0.217 & 0.159 & 0.774 & 0.665 & 0.285 \\
        ESMC 600M & 0.310 & 0.808 & 0.705 & 0.784 & \textbf{0.853} & \textbf{0.768} & \textbf{0.871} & 0.372 & 0.625 & 0.470 & 0.217 & 0.170 & 0.787 & 0.683 & 0.336 \\
        ProstT5 & 0.452 & 0.792 & 0.709 & 0.532 & 0.707 & 0.748 & 0.806 & 0.412 & 0.660 & 0.489 & 0.748 & 0.747 & \textbf{0.821} & \textbf{0.733} & 0.357 \\
        ProtT5 & 0.460 & 0.752 & 0.695 & \textbf{0.835} & 0.849 & 0.761 & 0.850 & 0.430 & 0.694 & 0.543 & 0.704 & 0.719 & 0.783 & 0.673 & \textbf{0.362} \\
        ProGen2 Small & 0.537 & \textbf{0.854} & 0.657 & 0.576 & 0.691 & 0.717 & 0.739 & 0.347 & 0.630 & 0.462 & 0.660 & 0.680 & 0.471 & 0.383 & 0.183 \\
        ProGen2 Medium & 0.517 & 0.825 & 0.674 & 0.724 & 0.756 & 0.733 & 0.792 & 0.392 & 0.667 & 0.512 & 0.719 & 0.732 & 0.567 & 0.464 & 0.217 \\
        ProGen2 Large & \textbf{0.796} & 0.563 & 0.702 & 0.754 & 0.762 & 0.744 & 0.832 & \textbf{0.432} & 0.649 & 0.559 & 0.709 & 0.729 & -- & -- & -- \\
        ProtGPT2 & 0.507 & 0.831 & 0.603 & 0.671 & 0.724 & 0.706 & 0.746 & 0.395 & 0.648 & 0.501 & 0.673 & 0.710 & -- & -- & -- \\ \bottomrule
    \end{tabular}%
    }
\end{table*}

\begin{table*}[th!]
    \centering
    \caption{Comparison of the performance of $k$-NN probes of the best layers of all PLMs and all tasks. Regression tasks (marked $^\dagger$) report Pearson correlation (PCC); all other tasks report Matthews correlation coefficient (MCC). ESM2-3B and ProGen2-Large required too much storage to compute residue-wise embeddings; ProtGPT2 has no per-residue embeddings, thus having no results for those tasks. }
    \resizebox{\textwidth}{!}{%
    \begin{tabular}{lcccccccccccc|ccc}
        & \multicolumn{12}{c|}{\textsc{protein-level tasks}} & \multicolumn{3}{c}{\textsc{residue-level tasks}}\\
        \textbf{Models} & \multicolumn{2}{c}{\textbf{Fluorescence}} & \textbf{GB1}$^\dagger$ & \multicolumn{2}{c}{\textbf{Meltome Atlas}} & \multicolumn{2}{c}{\textbf{Stability}} & \textbf{DeepSol} & \multicolumn{2}{c}{\textbf{DeepLoc2.0}} & \multicolumn{2}{c|}{\textbf{SCOPe40}} & \multicolumn{2}{c}{\textbf{SSP}} & \textbf{Binding} \\
        & Bin. & Reg.$^\dagger$ &  & $T_\text{m}^\dagger$ & Species & Rocklin$^\dagger$ & Tsubo.$^\dagger$ &  & Bin. & 10c & Fold & SF. & 3c & 8c & \\ \midrule
        ESM-2 8M & 0.583 & 0.690 & 0.850 & 0.434 & 0.742 & 0.726 & 0.890 & 0.294 & 0.657 & 0.488 & 0.693 & 0.753 & 0.649 & 0.534 & 0.475 \\
        ESM-2 35M & 0.614 & 0.711 & 0.854 & 0.472 & 0.741 & 0.740 & 0.905 & 0.320 & 0.675 & 0.537 & 0.796 & 0.867 & 0.703 & 0.585 & 0.475 \\
        ESM-2 150M & 0.613 & 0.715 & 0.866 & 0.573 & 0.788 & 0.744 & 0.911 & 0.369 & 0.689 & 0.578 & \textbf{0.837} & \textbf{0.901} & 0.737 & 0.621 & 0.489 \\
        ESM-2 650M & 0.621 & 0.722 & 0.879 & 0.622 & 0.799 & 0.747 & 0.911 & 0.377 & 0.695 & 0.580 & 0.834 & 0.893 & 0.753 & 0.640 & 0.515 \\
        ESM-2 3B & 0.619 & 0.717 & \textbf{0.904} & 0.628 & 0.791 & 0.746 & 0.911 & 0.370 & 0.697 & 0.563 & 0.817 & 0.880 & -- & -- & -- \\
        ESMC 300M & 0.647 & 0.735 & 0.867 & 0.704 & 0.851 & 0.749 & 0.923 & 0.366 & 0.700 & 0.576 & 0.817 & 0.870 & 0.745 & 0.626 & 0.466 \\
        ESMC 600M & 0.637 & 0.724 & 0.873 & \textbf{0.750} & \textbf{0.860} & 0.752 & \textbf{0.927} & 0.363 & 0.693 & 0.576 & 0.806 & 0.881 & 0.749 & 0.629 & 0.464 \\
        ProstT5 & 0.630 & 0.721 & 0.886 & 0.446 & 0.775 & 0.753 & 0.914 & 0.350 & 0.682 & 0.537 & 0.789 & 0.860 & \textbf{0.825} & \textbf{0.727} & \textbf{0.523} \\
        ProtT5 & 0.621 & 0.714 & 0.872 & 0.711 & 0.833 & \textbf{0.756} & 0.923 & \textbf{0.406} & \textbf{0.702} & \textbf{0.582} & 0.795 & 0.871 & 0.760 & 0.641 & 0.475 \\
        ProGen2 Small & 0.651 & 0.738 & 0.678 & 0.464 & 0.773 & 0.718 & 0.895 & 0.321 & 0.651 & 0.488 & 0.673 & 0.737 & 0.487 & 0.407 & 0.339 \\
        ProGen2 Medium & 0.652 & 0.739 & 0.776 & 0.587 & 0.771 & 0.731 & 0.899 & 0.371 & 0.660 & 0.515 & 0.729 & 0.802 & 0.555 & 0.464 & 0.365 \\
        ProGen2 Large & \textbf{0.741} & \textbf{0.791} & 0.746 & 0.637 & 0.793 & 0.725 & 0.897 & 0.363 & 0.665 & 0.541 & 0.683 & 0.749 & -- & -- & -- \\
        ProtGPT2 & 0.601 & 0.681 & 0.785 & 0.499 & 0.751 & 0.691 & 0.866 & 0.332 & 0.656 & 0.525 & 0.717 & 0.785 & -- & -- & -- \\ \bottomrule
    \end{tabular}%
    }
\end{table*}

\begin{table*}[th!]
    \centering
    \caption{Comparison of the performance of $k$-NN probes of the last layers of all PLMs and all tasks. Regression tasks (marked $^\dagger$) report Pearson correlation (PCC); all other tasks report Matthews correlation coefficient (MCC). ESM2-3B and ProGen2-Large required too much storage to compute residue-wise embeddings; ProtGPT2 has no per-residue embeddings, thus having no results for those tasks.}
    \resizebox{\textwidth}{!}{%
    \begin{tabular}{lcccccccccccc|ccc}
        & \multicolumn{12}{c|}{\textsc{protein-level tasks}} & \multicolumn{3}{c}{\textsc{residue-level tasks}}\\
        \textbf{Models} & \multicolumn{2}{c}{\textbf{Fluorescence}} & \textbf{GB1}$^\dagger$ & \multicolumn{2}{c}{\textbf{Meltome Atlas}} & \multicolumn{2}{c}{\textbf{Stability}} & \textbf{DeepSol} & \multicolumn{2}{c}{\textbf{DeepLoc2.0}} & \multicolumn{2}{c|}{\textbf{SCOPe40}} & \multicolumn{2}{c}{\textbf{SSP}} & \textbf{Binding} \\
        & Bin. & Reg.$^\dagger$ &  & $T_\text{m}^\dagger$ & Species & Rocklin$^\dagger$ & Tsubo.$^\dagger$ &  & Bin. & 10c & Fold & SF. & 3c & 8c & \\ \midrule
        ESM-2 8M & 0.539 & 0.648 & 0.764 & 0.434 & 0.695 & 0.705 & 0.881 & 0.281 & 0.631 & 0.470 & 0.647 & 0.710 & 0.649 & 0.534 & 0.475 \\
        ESM-2 35M & 0.540 & 0.638 & 0.723 & 0.472 & 0.724 & 0.718 & 0.896 & 0.320 & 0.623 & 0.509 & 0.693 & 0.747 & 0.703 & 0.583 & 0.433 \\
        ESM-2 150M & 0.541 & 0.633 & 0.798 & 0.573 & 0.788 & 0.715 & 0.910 & 0.338 & 0.650 & 0.555 & 0.589 & 0.636 & 0.724 & 0.610 & 0.401 \\
        ESM-2 650M & 0.511 & 0.626 & 0.813 & 0.622 & 0.799 & 0.684 & 0.906 & 0.377 & 0.633 & 0.529 & 0.560 & 0.605 & 0.736 & 0.624 & 0.412 \\
        ESM-2 3B & 0.417 & 0.559 & 0.876 & 0.628 & 0.791 & 0.661 & 0.895 & 0.360 & 0.649 & 0.533 & 0.595 & 0.656 & -- & -- & -- \\
        ESMC 300M & 0.544 & 0.648 & 0.822 & 0.692 & 0.851 & 0.729 & 0.919 & 0.338 & 0.610 & 0.499 & 0.344 & 0.354 & 0.689 & 0.580 & 0.267 \\
        ESMC 600M & 0.545 & 0.637 & 0.821 & \textbf{0.739} & \textbf{0.858} & 0.743 & 0.921 & 0.345 & 0.594 & 0.471 & 0.301 & 0.313 & 0.717 & 0.603 & 0.259 \\
        ProstT5 & 0.613 & 0.691 & \textbf{0.886} & 0.276 & 0.506 & 0.737 & 0.914 & 0.350 & 0.663 & 0.524 & \textbf{0.755} & \textbf{0.807} & \textbf{0.825} & \textbf{0.727} & \textbf{0.496} \\
        ProtT5 & 0.467 & 0.580 & 0.872 & 0.711 & 0.833 & \textbf{0.756} & \textbf{0.923} & \textbf{0.406} & \textbf{0.673} & \textbf{0.582} & 0.652 & 0.705 & 0.757 & 0.639 & 0.374 \\
        ProGen2 Small & 0.519 & 0.613 & 0.624 & 0.371 & 0.679 & 0.662 & 0.851 & 0.290 & 0.554 & 0.388 & 0.445 & 0.479 & 0.387 & 0.320 & 0.147 \\
        ProGen2 Medium & 0.551 & 0.621 & 0.598 & 0.538 & 0.742 & 0.654 & 0.847 & 0.328 & 0.560 & 0.382 & 0.435 & 0.469 & 0.386 & 0.319 & 0.131 \\
        ProGen2 Large & \textbf{0.715} & \textbf{0.776} & 0.661 & 0.634 & 0.793 & 0.714 & 0.890 & 0.344 & 0.612 & 0.478 & 0.518 & 0.545 & -- & -- & -- \\
        ProtGPT2 & 0.567 & 0.653 & 0.632 & 0.499 & 0.751 & 0.631 & 0.846 & 0.303 & 0.622 & 0.510 & 0.604 & 0.660 & -- & -- & -- \\ \bottomrule
    \end{tabular}%
    }
\end{table*}

\begin{figure*}[t!]
    \centering
    \includegraphics[width=\textwidth]{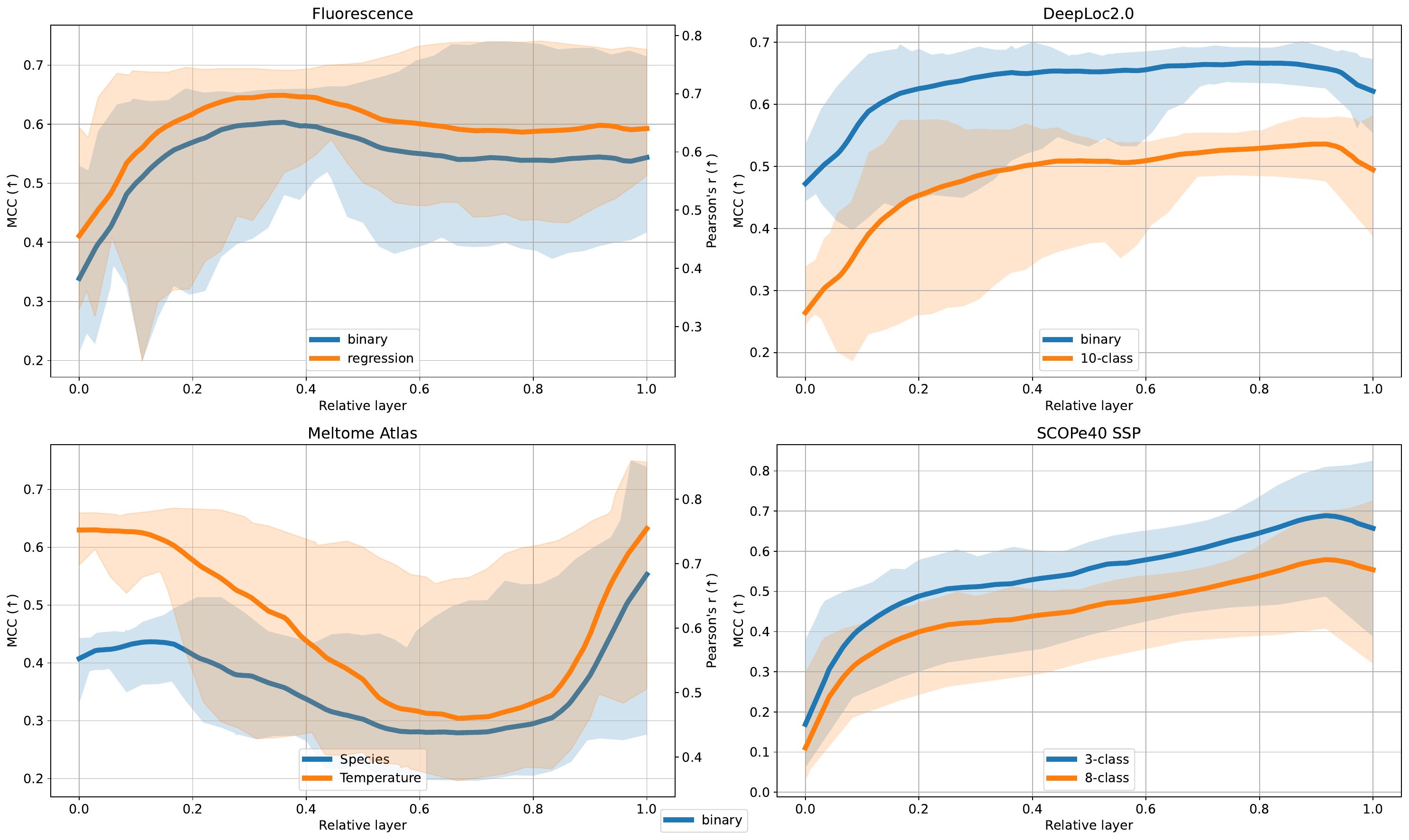}
    \caption{\textbf{$k$-NN ablation studies.} Pairwise comparison of two tasks for one dataset, showing that model performance patterns do not change across DTs within the same model and dataset. The Pearson correlation coefficients for these probes are $0.992\pm 0.006$ for fluorescence, $0.917\pm 0.079$ for the Meltome Atlas, $0.947\pm 0.036$ for DeepLoc2.0, and $0.999\pm 0.000$ for the SSP tasks on SCOPe40 2.08.}
\end{figure*}

\begin{figure*}[t!]
    \centering
    \includegraphics[width=\textwidth]{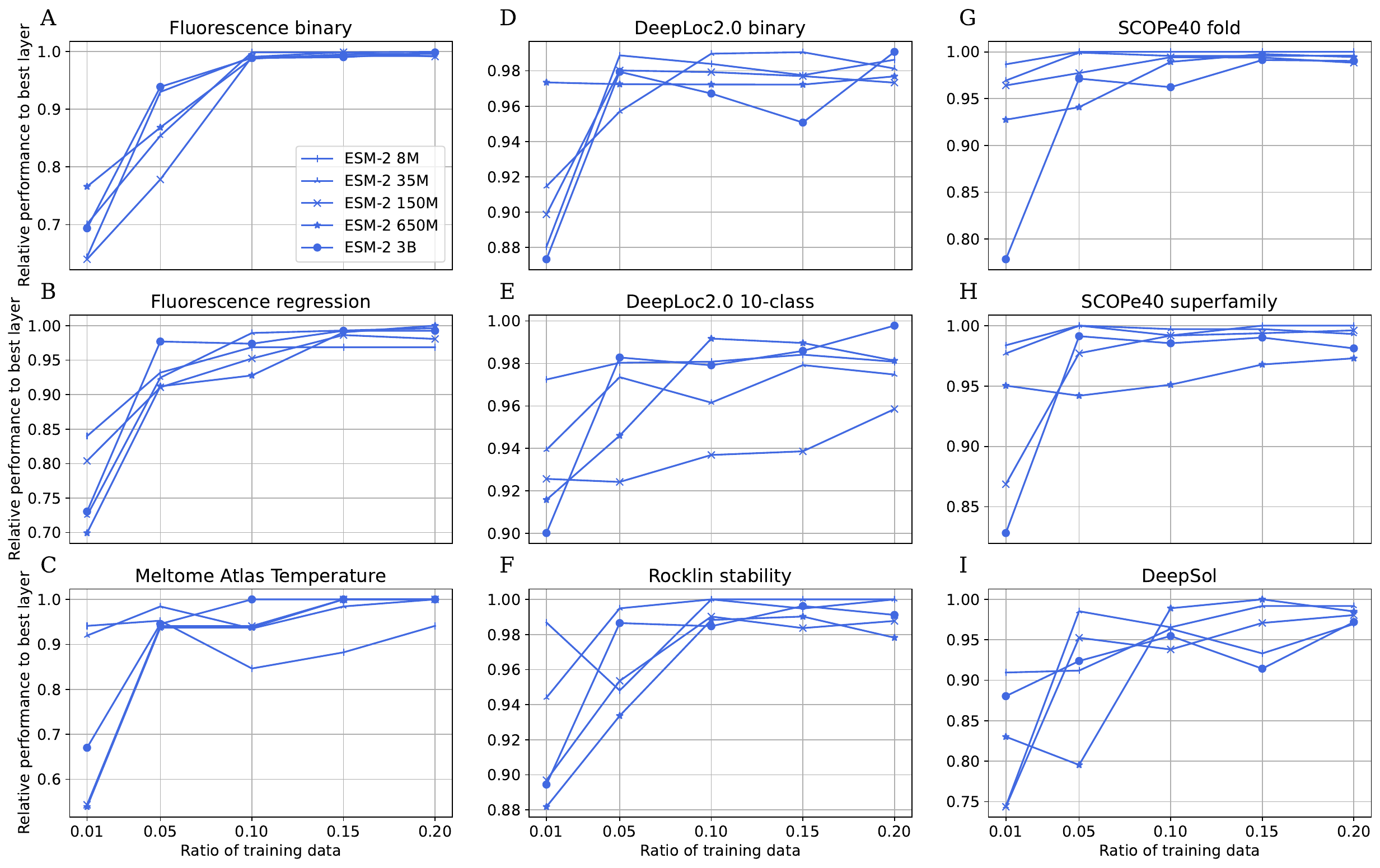}
    \caption{\textbf{Full sparse performance analysis.} Providing additional datasets to the four from the main text.}
\end{figure*}

\begin{figure}
    \centering
    \includegraphics[width=\textwidth]{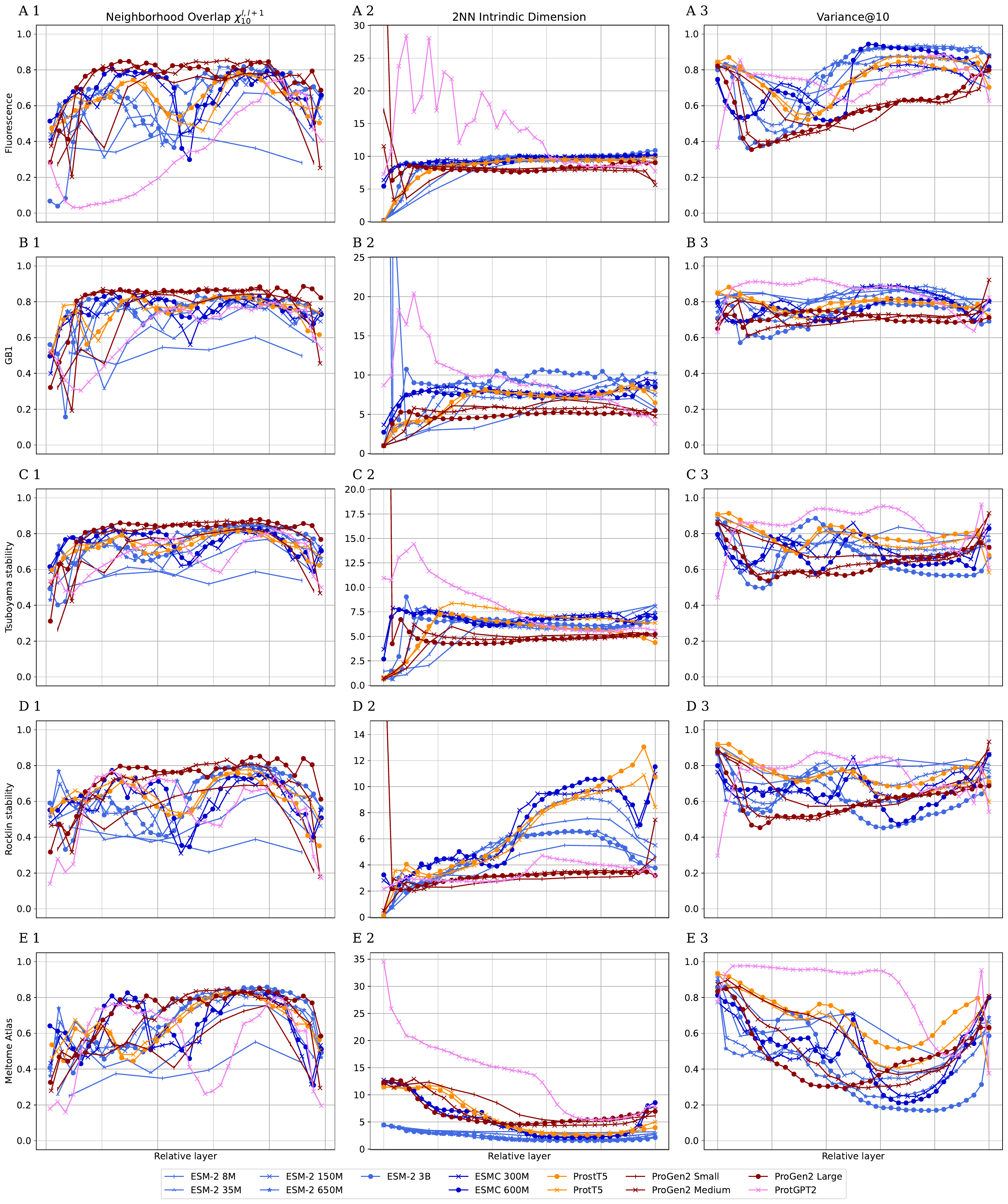}
    \caption{\textbf{Full latent space metrics.} Full model resolution of Figure 2.}
\end{figure}

\begin{figure}
    \centering
    \includegraphics[width=\textwidth]{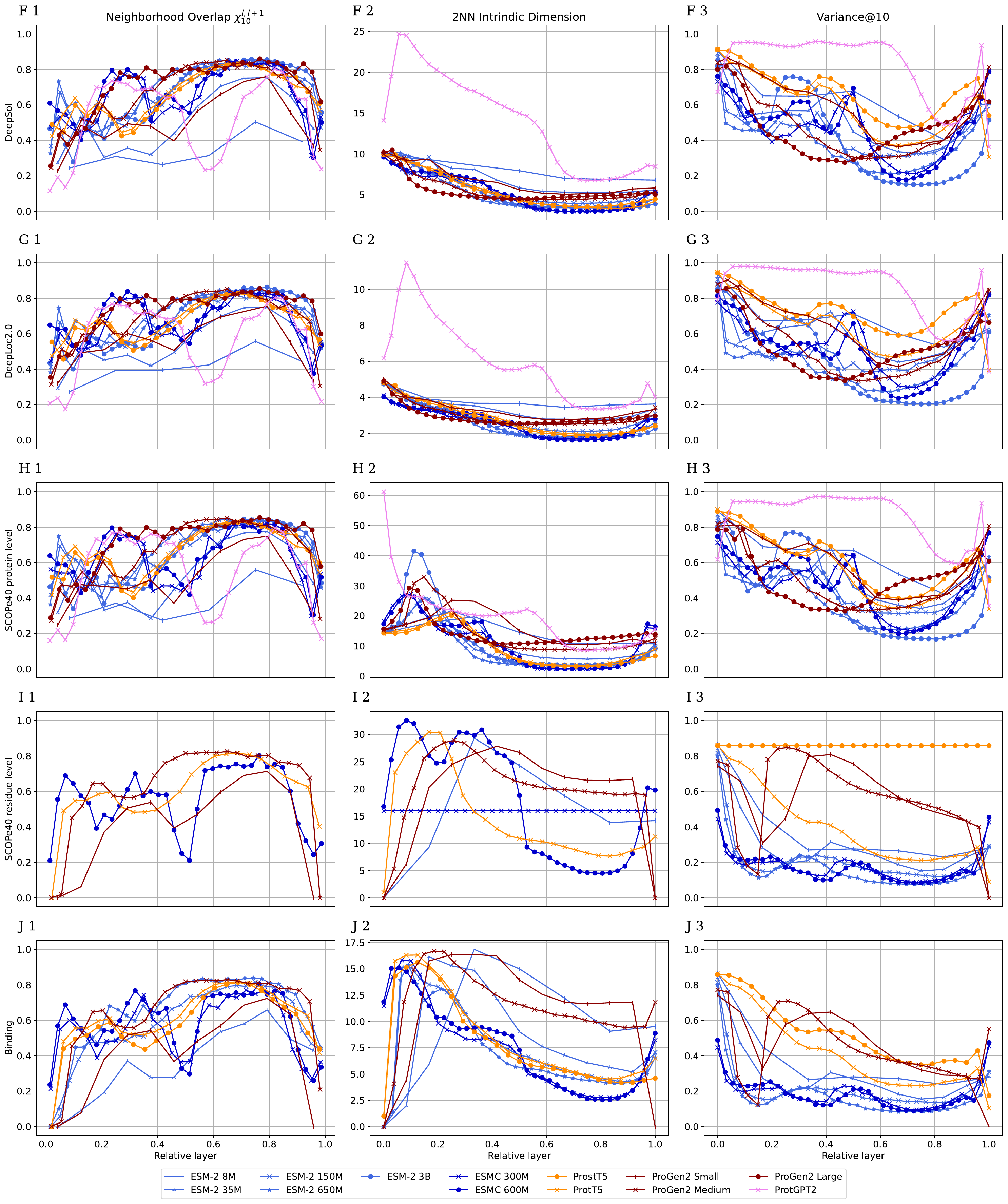}
    \caption{\textbf{Full latent space metrics.} Full model resolution of Figure 2.}
\end{figure}

\begin{table*}
    \centering
    \caption{\textbf{Datasets considered out-of-distribution compared to pre-training data.}}
    \begin{tabular}{lrr}\toprule
        Dataset & \# Sequences & Sequence Identity to UniProt \\ \midrule
        NovelMetaG~\cite{prabakaran2025deciphering} & 11,444 & $<0.3$ \\
        Lysozymes~\cite{madani2023large} & 190 & $0.71 \pm 0.13$ \\
        Chorismate mutases~\cite{russ2020evolution} & 2,748 & $0.70 \pm 0.14$ \\\bottomrule
    \end{tabular}
\end{table*}
\fi

\end{document}